\RequirePackage{fix-cm}
\documentclass[twocolumn]{svjour3}          
\smartqed  
\usepackage{graphicx}
\usepackage[utf8]{inputenc}
\usepackage{supertabular}
\usepackage{tabularx}
\usepackage{booktabs}

\usepackage{pdflscape}
\usepackage{fancyhdr} 
\fancypagestyle{mylandscape}{
\fancyhf{} 
\fancyfoot{
\makebox[\textwidth][r]{
  \rlap{\hspace{1cm}
    \smash{
      \raisebox{0.2cm}{
        \rotatebox{90}{\thepage}}}}}}
}
\usepackage{longtable}

\usepackage{array}
\usepackage{ragged2e}

\usepackage{subfigure}
\usepackage{amsmath}
\usepackage{xcolor}
\usepackage{color, soul}
\usepackage{array}
\usepackage{url}
\usepackage{hyperref}

\begin{document}

\title{Robot's Gendering Trouble: A Scoping Review of Gendering Humanoid Robots and its Effects on HRI 
}


\author{Giulia Perugia         \and
        Dominika Lisy
}


\institute{G. Perugia, Human-Technology Interaction Group, Eindhoven University of Technology, Eindhoven, Netherlands              
             (\email{g.perugia@tue.nl})
             \\
D. Lisy, Department of Thematic Studies, Division of Gender Studies, Linköping University (E-mail: dominika.lisy@liu.se)  
}

\date{Received: date / Accepted: date}

\maketitle

\begin{abstract}
The discussion around gendering humanoid robots has gained more traction in the last few years. To lay the basis for a thorough understanding of how robots' ``gender" has been understood within the Human-Robot Interaction (HRI) community -- i.e., how it has been manipulated, in which contexts, and which effects it has yielded on people's perceptions and interactions with robots -- we performed a scoping review of the literature. We identified 553 papers relevant for our review retrieved from 5 different databases. The final sample of reviewed papers included 35 papers written between 2005 and 2021, which involved a total of 3902 participants.
In this article, we thoroughly summarize these papers by reporting information about their objectives and assumptions on gender (i.e., definitions and reasons to manipulate gender), their manipulation of robots' ``gender" (i.e., gender cues and manipulation checks), their experimental designs (e.g., demographics of participants, employed robots), and their results (i.e., main and interaction effects). The review reveals that robots' ``gender" does not affect crucial constructs for the HRI, such as likability and acceptance, but rather bears its strongest effect on stereotyping. We leverage our different epistemological backgrounds in Social Robotics and Gender Studies to provide a comprehensive interdisciplinary perspective on the results of the review and suggest ways to move forward in the field of HRI.
\end{abstract}

\section{Introduction}

Gender studies emerged as an academic discipline in the 1980s to study and understand the nuances of how gender is imbued in the power structures of society, as well as how gender materializes in the design of objects, spaces, and knowledge practices \cite{lykke2010feminist}. Gendered design is common in machines and objects \cite{ehrnberger2012visualising}, for instance, in medical devices \cite{ehrnberger2017androchair,johnson2008simulating} as well as children's toys \cite{fine2018does,reich2018constructing}, and is oftentimes deemeed necessary to accommodate individual differences and users' preferences \cite{moss2006some}. More often than not, however, gendered design is redundant and conducive of stereotypes and binary perspectives on gender (i.e., the understanding that gender includes only two discrete and opposite categories of female and male \cite{butler1999gendertrouble}) \cite{de2022inclusive}. The inherent binarism of gender has been heavily contested with the emergence of feminist and queer theory for its normative power and exclusionary potential \cite{butler1999gendertrouble,lykke2010feminist}. Gendered robots are a particularly interesting case of gendered design as their ``gender'' often derives from their humanoid shape, and is thus deeply entangled with the human body \cite{perugia2021gender,perugia2022shape}. There is still little knowledge about what exactly it means to “gender” a humanoid robot and how the gendering of robots impacts users' perception and interaction with them.
 In this scoping review, we are particularly interested in the emergence of the practice of gendering humanoid robots in Human-Robot Interaction (HRI) research to assess its feasibility and consequences and identify ways to move forward.

\subsection{A Perspective from Gender Studies}
``What \textit{is} gender?'' seems to be the imperative question with regards to gendered robots which presupposes the idea that gender \textit{is} a concrete thing. In feminist theory and the academic field of Gender Studies, the object of study is assumed to be ``gender'' (see \cite{butler1994against,lykke2010feminist}), yet the interest does not lie in identifying the essence of gender as a fixed category but rather in recognizing the transformative value of gender as a system of thought and a practice. 
Once gender is not anymore understood as an inherent characteristic or physical attribute of a body but instead as an organizing principle embedded in social structures, behavior, design, and norms, it can be seen as a \textit{lens} that organizes human life and the knowledge about human bodies.  
Thus, assessing the effect of ``gender'' in robots through the theoretical lens of Gender Studies shifts the emphasis from gender as a fixed property of robot bodies to the investigation of gendering practices of robot development and testing.

Historically, the distinction between sex and gender (or lack thereof) has been influential for acknowledging the socio-culturally constructed aspects of being a woman or being a man in the wider society and the roles attached to it.  
The fact that gender is assumed to derive from sex strengthens the idea of an essential difference between men and women \cite{lykke2010feminist,butler1994against}.  
Prominent feminist philosopher Butler \cite{butler1999gendertrouble} introduced the false dichotomy of sex and gender, and argued that sex is as equally socially constructed as gender. Through this argument, Butler emphasized the performativity of gender (i.e. a repetitive, ritualized process of talking about and doing gender as a social act \cite{butler1988performative}) and its use as a principle to organize human bodies and knowledge. Moving from thinking of gender as an attribute (``having a sex/gender'') or an essence (``being a sex/gender'') to thinking of it as an organizing principle allows a theoretical shift from the analysis of gender as a social marker to the analysis of \textit{gendering} as a process (how ``gender’’ is \textit{done}) \cite{butler1999gendertrouble}. Beginning to trouble what ``gender" means for robot design and attempting to focus on how ``gender" is \textit{done} by roboticists is at the core of this review.

In most cases, gendering is a process of dividing into two categories and hierarchically positioning them in opposition to one another \cite{lugones2007heterosexualism}\cite{lugones2010toward}. If an object is conceived as masculine, it is associated with concepts opposed to femininity. This is not necessarily problematic but can be problematic when designers are oblivious to the hierarchy imbued in these gendered categorizations and the resulting social consequences of certain design choices \cite{alesich2017gendered}. Gendering humanoid robots means mapping them onto the gendering of human bodies and their hierarchical positioning and other intersected structures of power \cite{dignazio2020data}.
This entails that the design of this technology is inherently political and likely to reinforce power structures and hierarchies of domination \cite{wajcman2004technofeminism,balsamo1996technologies,faulkner2001technology,dignazio2020data}. In addition, the under-representation of women and other marginalized identities in the development of technology contributes to these power imbalances
 (see \cite{cockburn1992circuit,dignazio2020data}).

Feminist theory urges to shift from a rather uncritical engagement with technology design and testing to acknowledging the transformative and relational potential of technology. 
If gender continues to be treated uncritically in relation to technology, the danger is, as Balsamo puts it, that ``new technologies will be used primarily to tell old stories - stories that reproduce, in high-tech guise, traditional narratives about the gendered, race-marked body" \cite{balsamo1996technologies}. Through a critical engagement, feminist theory developed modes of inquiry into the gendered knowledges and practices and intersectional structures of power \cite{lykke2010feminist,dignazio2020data}.
A deeper engagement with ideas and practices of gendering robots from the Feminist and Gender Studies scholarship would likely exceed the scope of this literature review. With this section, we wanted to introduce core ideas from Gender Studies that could illuminate the results of this review and provide the HRI scholarship with a different, more complex, understanding of the concept of ``gender." We acknowledge the many epistemological differences between the two fields of studies, but nevertheless hope to inspire an interdisciplinary cross-pollination that could enrich the understanding of what is at stake with regards to the \textit{gendering} of robots. 

\subsection{Gender in Robotics}

Currently, there is still little knowledge about the effects of gendering robots and what exactly it entails to ``gender'' a robot. This begs the question whether ``gender'' can be a useful or harmful design feature in humanoid robots. 
``Gender'' as a design variable and structuring element in robotics is a relatively emergent field of inquiry with only a few theoretical engagements. The need to address the issue of gendering practices in robotics developed through critical analysis of prevalent bias towards high-pitched voice assistants on the market, which have been criticized for promoting stereotypes in gendered job associations and normalization of abuse against women \cite{west2019blush,alesich2017gendered,loideain2020alexa}. 
With the increase of robotic technologies used in social settings, aspects like the gendered voice and embodiment of the robot are inevitably in need of critical examination. Thus, testing for a preference of gendered robots is receiving increased attention. 

Within the robotics community only a few scholars have contributed to the theoretical discussion about the role of gender and asked for a more elaborate and sensitive investigation. According to Nomura \cite{nomura2017robots}, the influence of gender markers in interactions between humans immediately suggests the relevance of gender cues in interaction with robots. However, Nomura highlights that the context and quality of the interaction might be more prevalent than gender itself in influencing people's perception of the interaction with the robot. Most importantly, the need for gendering and its ethical implications (i.e., confirming gender role stereotypes) is at the heart of Nomura’s critique. He emphasizes the need for a deeper discussion on the topic of implementing gendered features in robots. In line with Nomura, Alesich and Rigby \cite{alesich2017gendered} argue that there is still a lack of knowledge about the effect of gendering robot design. Roboticists are often not aware of the interweavings of gender and human bodies and how it organizes society and values. The focus on technical problem solving and the fast-paced testing and production in research and industry do not allow for ethical considerations of the social consequences that implementing ``gender" in robot design would require \cite{alesich2017gendered}. 
Thus, critically engaging with gendering practices in HRI is highly recommended.
 
S\o raa \cite{soraa2017mechanical} introduces the idea of mechanical genders for robots, which mirror the physical and social aspects of human gender as understood in the field of psychology (which commonly distinguishes between biological, social and  psychological gender).
S\o raa's theorization acknowledges the invented and mirroring effect of modeling robot ``gender" after human gender while preserving the difference between them. Most importantly, S\o raa \cite{soraa2017mechanical} highlights the bidirectional nature of gendering and argues that humanoid robots cannot be ``genderless''. Indeed,  roboticists' and users' understanding and ideas about humans as a category  are inevitably influenced by a gendered perspective and likely to flow into the design or perception of humanoid robots. This suggests that gendering might not be an entirely controllable process.

The need and interest to address gendering practices in robotics is evident.
Interdisciplinary work is still lacking in this regard, and this review attempts an \textit{interdisciplinary overview and analysis of robot's ``gender" that integrates the different epistemological traditions of Social Robotics and Gender Studies} to address whether imbuing robots with gender cues is a viable and ethical design direction for HRI. 

\subsection{Positionality and Terminology}
\label{positionality}

In approaching this review, we want to be transparent in our personal positioning and critical approach towards the concept of gender and its use in experiments. As women, we are affected personally by potential stereotyping effects of gendered robot design and so we have our stakes in gaining a nuanced understanding and a productive, yet sensitive, way forward in future research practices. This is in no way clouding our ability to assess and reason about advantages and disadvantages of gendering practices. 
Since a lot of the reviewed studies referred to gendered robots as female and male, we kept the same terms in our writing. This is primarily a way to circumvent confusion and elucidate the terminology used in these papers. However, in this article, we try to shift the thinking towards the process of ``gendering'' a robot and the ``genderedness'' of a robot, both described by Perugia et al. \cite{perugia2022shape}. According to Perugia et al. \cite{perugia2022shape}, the process of \textit{gendering} a robot is a two-step process of gender \textit{encoding}, in which designers imbue robots with gendered cues, and gender \textit{decoding}, in which users attribute ``gender" to robots. Gender encoding is an optional step, which can be avoided by resorting to robots with minimal anthropomorphic cues or minimized by avoiding adding gender cues to already gendered robot embodiments. Gender decoding, instead, seems to be a spontaneous process. Indeed, it occurs when designers imbue robots with gender cues but also when they do not, as shown, among others, by Marchetti-Bowick in their work on the attribution of gender to the Roomba vacuum cleaner \cite{marchetti2009your}. The present scoping review focuses on the encoding phase of the gendering process, how it is performed by the HRI scholarship, and the effect it has on the HRI. We touch upon gender decoding only when discussing the robot's manipulation check.

In performing this review, we adopt the epistemological perspective of Social Robotics, both in terms of methods and in terms of object of inquiry (i.e., the experimental manipulation of robot's genderedness). Taking a more experimental and techno-centric approach entails consistently simplifying the discussion of gender with respect to its complexity as outlined in this Introduction. 
We integrate the lens of Feminist and Gender Studies in the discussion to outline and highlight the potential implications of current HRI research practices. 
In the following sections, we describe the core objectives and research questions of our scoping review (see Section \ref{objectives}), detail the method we used to retrieve the papers included in the review (see Section \ref{methods}), report the findings of the reviewed papers (see Section \ref{results}), and critically examine these findings in our discussion with the aim of coming up with guidelines on how to move forward in the field of HRI (see Section \ref{discussion}).

\section{Objectives \& Research Questions}
\label{objectives}

The goal of this scoping review is to describe how the HRI scholarship has understood and manipulated ``gender" in humanoid robots, summarize the effects of robot's genderedness on the perception of and interaction with humanoid robots, and identify best practices to manipulate a robot's genderedness from a feminist perspective. In parallel with these main objectives, this scoping review also aims to appraise the reason for manipulating the robot's genderedness and the validity of such manipulation.  
We attempt to answer the following research questions (RQ):

\begin{itemize}
\item \textbf{RQ1.} \textit{How has the robot's genderedness been manipulated by the HRI scholarship?}
\item \textbf{RQ2.} \textit{What role does the robot's genderedness play in the perception and interaction with humanoid robots?}
\end{itemize}

\begin{figure*}
    \centering
    \includegraphics[width=0.75\textwidth]{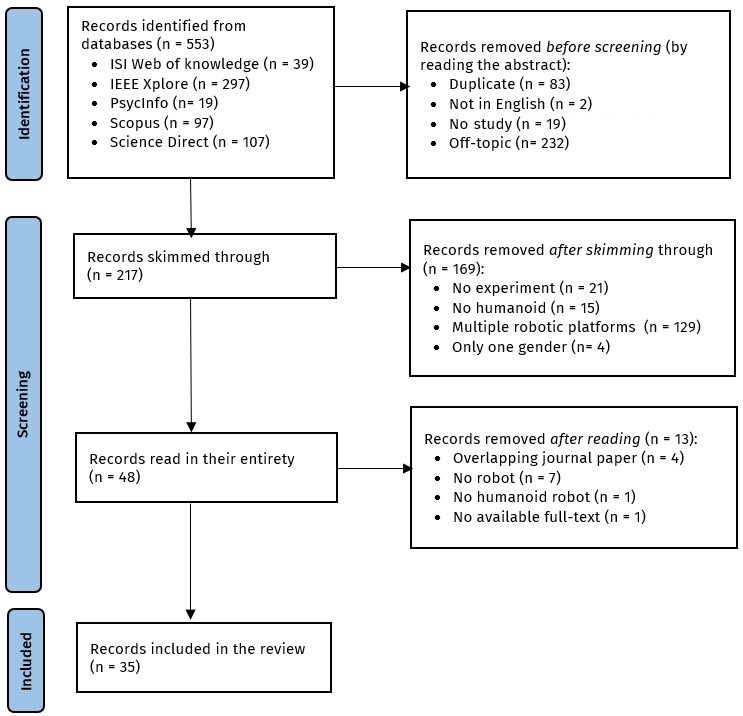}
    \caption{\textbf{PRISMA diagram detailing the paper selection pipeline}}
    \label{fig:selection_pipeline}
\end{figure*}

\section{Methodology}
\label{methods}

\subsection{Data Collection \& Eligibility Criteria}
\label{datacollection}

In order to identify the papers to include in this scoping review, we performed an electronic search in the following databases: IEEE Xplore, Scopus, ISI Web of Science (WoS), PsycINFO, and Science Direct. We used the following three variations of the same search string. The variation depended on the number of wildcards (*) that each database accepted:
\begin{itemize}
    \item [1.] ``robot gender*'' OR ``gender of robot*'' OR ``gender of the robot*'' OR ``gender* robot*'' OR ``male* robot'' OR ``female* robot'' OR (``gender cue*'' AND ``robot*'')
    \item [2.] ``robot gender*'' OR ``gender of robot'' OR ``gender of the robot'' OR ``gender* robot*'' OR ``male* robot'' OR ``female* robot'' OR (``gender cue*'' AND ``robot*'') 
    \item [3.] ``robot gender'' OR ``gender of robot'' OR ``gender of the robot'' OR ``gender robot'' OR ``male robot'' OR ``female robot'' OR (``gender cue'' AND ``robot'')
\end{itemize}
The search was performed independently by the two authors. GP focused on ISI Web of Science and Science Direct, whereas DL on IEEE Xplore, PsycInfo, and Scopus. The search yielded a list of 553 papers (May 2021) of which:
\begin{itemize}
    \item 39 from ISI Web of Science (search string 1)
    \item 297 from IEEE Xplore (search string 2)
    \item 19 from PsycInfo (search string 1)
    \item 97 from Scopus (search string 1)
    \item 107 from Science Direct (search string 3)
\end{itemize}

The papers obtained from the electronic search were imported in a shared spreadsheet and screened against the following eligibility criteria: (i) the papers were written in English,  (ii) included the manipulation of at least two ``genders" of the robot (e.g., studies including only female robots were excluded), (iii) manipulated the robot's genderedness through the same robotic platform (e.g., studies manipulating two ``genders" but with different robotic platforms were excluded), (iv) focused on physical humanoid robots or virtual instantiations of humanoid robots, (v) did not focus on sex robots, and (vi) reported experimental results. These exclusion and inclusion criteria were set so that we could easily identify the cues that the HRI scholarship resorted to to modify the robot's genderedness.
The inclusion of papers focusing only on one ``gender" or manipulating genderedness with different robotic platforms would have not allowed us to isolate these cues so easily as other factors, such as differences in the robots' embodiments, materials, body parts, humanlikeness, could have influenced the researchers' choice of the cues to use. In the next section, we describe the three steps of the selection pipeline process in more detail. 

\subsubsection{Selection Pipeline}
\label{selection_pipeline}

From the initial batch of 553 papers, we removed duplicate results, front covers, and tables of contents. This process left us with 470 papers (see Figure \ref{fig:selection_pipeline} for the diagram of the selection pipeline). We read the abstracts of all 470 papers and excluded 253 papers that were not in English ($N=2$), did not present an experimental study (e.g., theoretical paper) ($N=19$), or were off-topic ($N=232$). This process resulted in 217 papers.

In a second exclusion round, we skimmed through the papers' content and excluded 169 papers that did not feature any experiment or robot ($N=21$), did not include a humanoid robot ($N=15$), did not manipulate the genderedness of the robot or manipulated it but using multiple robotic platforms ($N=129$), and focused on just one ``gender" ($N=4$). After this step, we were left with 48 papers. 

These 48 papers were divided between the authors and read in their entirety. GP read 29 of the papers, DL 17. Of this batch of papers, 13 papers were excluded because they were short versions of a longer journal paper already featured in our list ($N=4$), did not employ a robot ($N=7$), employed a robot that was not humanoid ($N=1$), or did not have a full-text available online ($N=1$). As a result of the selection pipeline, we included 35 papers written between 2005 and 2021 in our scoping review. Out of these 35 papers, 7 were journal papers, 17 were full papers included in the proceedings of a conference,
10 were short papers included in the proceedings of a conference, and 1 was a workshop paper. 
The selection process is described in Figure \ref{fig:selection_pipeline}. The last search was performed in May 2021.

\subsection{Coding \& Information Extraction}
\label{coding}
Once obtained the final list of 35 papers to include in our scoping review, we performed a thorough work of coding and information extraction. For each paper, we recorded:
\begin{enumerate}
    \item \textit{General information}: the name of the authors, the year of publication of the paper, and the type of paper (i.e., conference or journal, short or full paper; see Section \ref{selection_pipeline}).
    \item \textit{Experimental information}: the number of participants in the study, their age and gender, the robot used in the study, the type of embodiment of the robot (e.g., picture, video, physical), the independent variables (beyond the robot's genderedness), the dependent variables, and the type of task used in the study (see Tables \ref{table:General and demographic information} and \ref{table:Experimental_Information}, and Section \ref{taskandactivities}).
    \item \textit{Gender-related information}: definitions of gender, reasons to manipulate the robot's genderedness in the first place, ``genders" manipulated (e.g., female, male and gender neutral robots), cues used to manipulate the robots' genderedness, presence of a manipulation check, metrics used to perform the manipulation check, and rationale behind the choice of the cues (see Table \ref{table:GenderManipulation1}, and Sections \ref{definitions_motivations}, \ref{gendermanipulation} and \ref{manipulationANDassessment}).
    \item \textit{Results}: main effects of the robot's genderedness and interaction effects of robot's genderedness and other independent variables on the dependent variables (See Table \ref{table:Experimental_Information} and Section \ref{effects}).
\end{enumerate}

Tables \ref{table:General and demographic information}, \ref{table:GenderManipulation1}, and \ref{table:Experimental_Information} report part of the results of the coding and information extraction process, as well as the summaries of all 35 papers. The rest of the extracted information is presented in the Results section.

\section{Results}
\label{results}

\subsection{Characteristics of the Included Studies}

\subsubsection{Participants}

\begin{figure*}
    \centering
    \includegraphics[width=.8\textwidth]{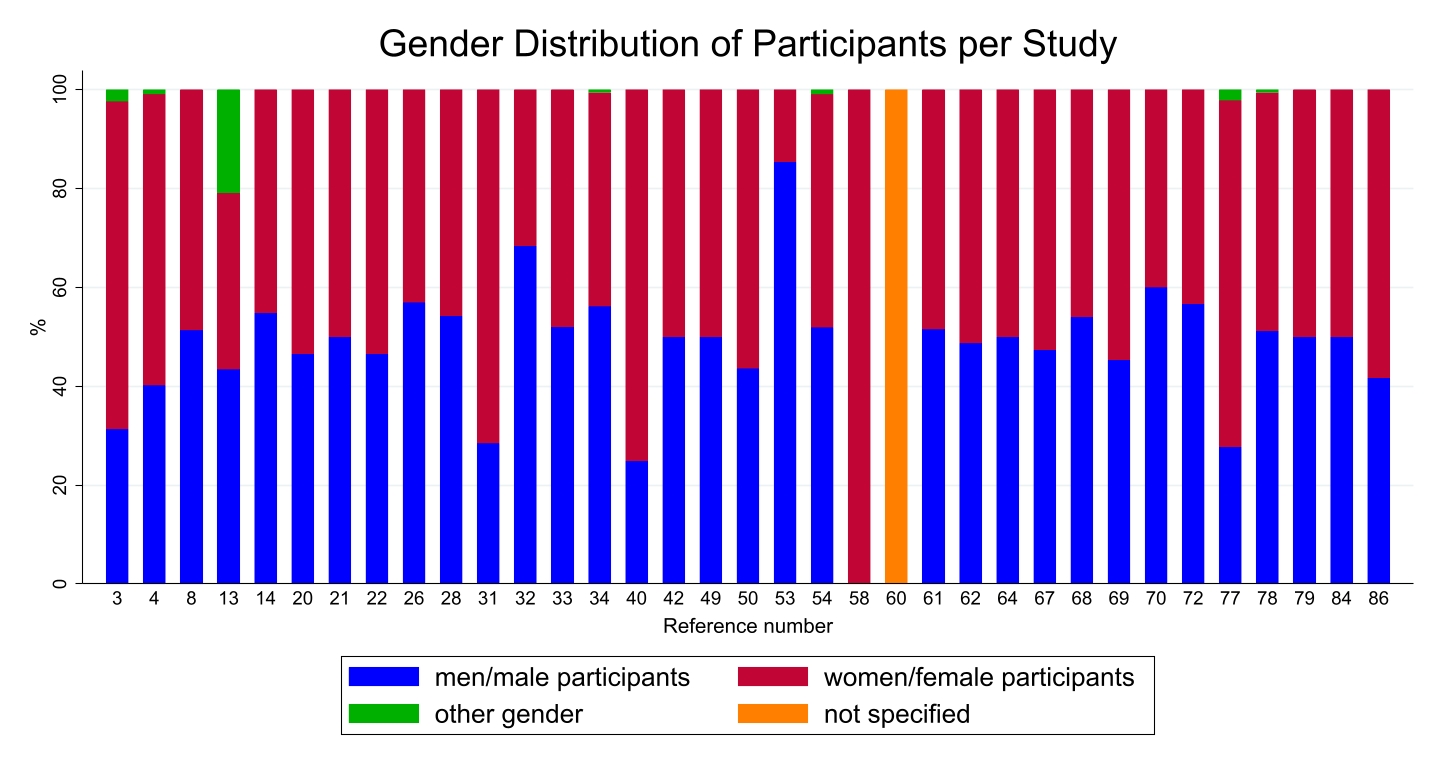}
    \caption{\textbf{Distribution of Participant’s Gender in the Reviewed Studies.} In blue, men/male participants, in red, women/female participants, in orange, participants whose gender was not specified and in green, participants falling into the other/undisclosed gender category.
}
    \label{fig:participantgender}
\end{figure*}

Overall, the studies reported in the papers included 3902 participants (see Table \ref{table:General and demographic information}).
The participants in the studies were more or less equally distributed between female (49\%) and male gender (47\%, see Figure \ref{fig:participantgender} for an overview). Interestingly, only 1\% of the participants in the studies fell in the category other/undisclosed, and the gender of 3\% of the participants was not specified. None of the reviewed studies reported the presence of non-binary participants or participants with gender identities beyond the binary. In terms of age, 60\% of the papers featured a sample of participants composed of young adults, presumably university students (age comprised between 18 and 30 years); 20\% of the papers a sample of adults (older than 30), and 20\% of the papers a sample of children (younger than 18).

\begin{table*}[]
\renewcommand{\arraystretch}{1.5}%
\centering
\caption{General and demographic information about the studies included in the scoping review (F= female, M= male, \textit{dns}= did not specify their gender). The terms used for participants' gender in the tables are derived from the papers. Studies \cite{eyssel2012if} and \cite{eyssel2012activating} refer to the same study but report the results of different dependent variables. *= no manipulation of gender in this study; $\approx$= calculated from partial means (when only group means are reported); \textit{(?)}= it is not clear from the paper whether participants interacted with a physical robot.  }
\begin{tabular}{ r | c c c  | c c }  &  \multicolumn{3}{c|}{\textbf{Participants}} & \multicolumn{2}{c}{\textbf{Robot}} \\
\textbf{Authors (date)} & \textbf{$N$} & \textbf{Age ($M$)} & \textbf{Gender} & \textbf{Name} & \textbf{Embodiment} \\
\hline
Bernotat et al. (2017) \cite{bernotat2017shape}  & 83 & 26.15 & 26M, 55F, 2\textit{dns} & Meka M1 & Pictures \\ 
Bernotat et al. (2021) \cite{bernotat2021fe} & 107 & 27.23 & 43M, 63F, 1\textit{dns} & Meka M1 & Pictures \\ 
Bryant et al. (2020) \cite{bryant2020should} & 150 & 40.02 & 51\%M, 49\%F & Pepper & Video \\ 
Calvo-Barajas et al. (2020) \cite{calvo2020effects}  & 129 & 11.29 & 56M, 46F, 27\textit{dns} & Furhat & Physical \\
Chita-Tegmark et al. (2019) \cite{chita2019gender} & 197 & 36.10 & 44\%F & Willow garage PR2 & Video \\
Chita-Tegmark et al. (2019) \cite{chita2019gender} & 197 & 35.12 & 47\%F & Willow garage PR2 & Video (vignette)\\
Chita-Tegmark et al. (2019) \cite{chita2019gender} & 100 & 33.73 & 44\%F & Willow garage PR2 & Video (vignette)* \\
Eyssel \& Hegel (2012) \cite{eyssel2012s}  & 60 & 24 & 30M, 30F  & Flobi & Pictures \\
Eyssel et al. (2012a) \cite{eyssel2012if}  & 58 & 22.98 & 27M, 31F & Flobi & Video \\
Eyssel et al. (2012b) \cite{eyssel2012activating} & 58 & 22.98 & 27M, 31F & Flobi & Video \\
Ghazali et al. (2018) \cite{ghazali2018effects} & 72 & 23.90 & 41M, 31F & Socibot & Physical \\
Jackson et al. (2020) \cite{jackson2020exploring} & 118 & 37.36 & 64M, 54F & NAO & Video \\
Jung et al. (2016) \cite{jung2016feminizing}& 144 & 20.21 & 103F & \textit{not specified} & Physical \\
Kraus et al. (2019) \cite{kraus2018effects} & 38 & 26.34 & 26M, 12F & NAO & Physical \textit{(?)} \\
Kuchenbrandt et al. (2014) \cite{kuchenbrandt2014keep} & 73 & 25.04 & 38M, 35F & NAO & Physical \\
Law et al. (2020) \cite{law2021interplay} & 198 & 34.96 & 95F, 1\textit{other} & Willow garage PR2 & Video (vignette)\\
Law et al. (2020) \cite{law2021interplay} & 421 & 36.52 & 162F, 3\textit{other} & Willow garage PR2 & Video (vignette) \\
Lugrin et al. (2020) \cite{lugrin2020if}  & 205 & 28.10 & 24.9\%M, 75.1\%F & Reeti & Video \\
Makenova et al. (2018) \cite{makenova2018exploring} & 36 & 34.3 & 18M, 18F & NAO & Physical \\
Nomura \& Kinoshita (2015) \cite{nomura2015gender} & 20 & 20.4 & 10M, 10F & Robovie-SX & Physical \\
Nomura \& Takagi (2011) \cite{nomura2011exploring} & 39 & \textit{not specified} & 17M, 22F & Robovie-X & Physical \\
Paetzel et al. (2016a) \cite{paetzel2016congruency} & 48 & 23.96 & 14.6\%F & Furhat & Physical \\
Paetzel et al. (2016b) \cite{paetzel2016effects}  & 106 & $\approx11.69$ & 55M, 50F, 1\textit{other} & Furhat & Physical \\
Pfeifer \& Lugrin (2018) \cite{pfeifer2018female}  & 45 & 20.51 & 45F & Reeti & Physical \\
Powers and Kiesler (2006) \cite{powers2006advisor} & 98 & \textit{not specified} & \textit{not specified} & \textit{not specified} & Video \\
Powers et al. (2005) \cite{powers2005eliciting} & 33 & 21 & 17M, 16F & \textit{not specified} & Physical \\
Rea et al. (2015) \cite{rea2015check} & 39 & \textit{not specified} & 19M, 20F & NAO & Physical \\
Reich-Stiebert \& Eyssel (2017) \cite{reich2017ir} & 120 & 24.57 & 60M, 60F & NAO & Physical \\
Sandygulova \& O'Hare (2018) \cite{sandygulova2018age} & 55 & \textit{not specified}  & 33M, 22F & NAO & Physical \\
Sandygulova \& O'Hare (2016) \cite{sandygulova2016investigating} & 74 & $\approx5.8$  & 40M, 34F & NAO & Physical \\
Sandygulova \& O'Hare (2015) \cite{sandygulova2015children} & 64 & \textit{not specified} & 29M, 35F & NAO & Physical \\
Sandygulova et al. (2014) \cite{sandygulova2014investigating} & 76 & \textit{not specified} & 36M, 40F & NAO & Physical \\
Siegel et al. (2009) \cite{siegel2009persuasive}& 134 & 35.6 & 76M, 58F & Nexi & Physical \\
Tay et al. (2014) \cite{tay2014stereotypes}& 164 & 35.6 & 84M, 79F, 1\textit{dns} & \textit{not specified} & Physical \\
Thellman et al. (2018) \cite{thellman2018he} & 118 & 22.47 & 59M, 59F & NAO & Physical \\
You \& Lin (2019) \cite{you2019gendered} & 64 & 24.2 & 32M, 32F & Alpha 1 Pro & Physical \\
Zhumabekova et al. (2018) \cite{zhumabekova2018exploring} & 24 & $\approx6.7$ & 10M, 14F & NAO & Physical \\
Steinhaeusser et al. (2021) \cite{steinhaeusser2021anthropomorphize} & 137 & 26.21 & 38M, 96F, 3\textit{other} & NAO & Video \\
\end{tabular}
\label{table:General and demographic information}
\end{table*}

\begin{table*}[]
\footnotesize
\centering
\caption{Manipulation of the robot's genderedness in the studies included in the scoping review: robot's ``genders" manipulated (M= male; F= female; N= neutral), cues used to manipulate the robot's genderedness, presence of a manipulation check (Yes= manipulation check is performed; No= manipulation check is not performed; \textit{ns}= no statistic performed to verify the manipulation check), significance of the manipulation check (\textbf{bold}=significant, \textit{italics}=partially significant), metrics used to assess perceived gender, and notes.}
\begin{tabular}{ l | l | l | l | l | l } \textbf{Authors} & \textbf{Robot's} & \textbf{Cues} & \textbf{Manip.} & \textbf{Metric} & \textbf{Notes} \\
 & \textbf{``Genders"} & \textbf{Used} & \textbf{Check} & \textbf{Used}\\[2ex]
\hline\noalign{\smallskip}
Bernotat & M, F & Body & \textbf{Yes} & Participants indicated the extent to &\\
et al. (2017) \cite{bernotat2017shape} &  & Proportion &  & which they would perceive the robots &\\
& &  &  & as male and female on 2 items (7-point & \\
& &  &  & Likert scale)&\\[2ex]
Bernotat & M, F  & Body & \textbf{Yes} & Participants indicated the extent to &\\
et al. (2021) \cite{bernotat2021fe} &  & Proportion &  & which they would perceive the robots \\
& &  &  & as male and female on 2 items (7-point & \\
& &  &  & Likert scale)&\\[2ex]
Bryant  \cite{bryant2020should} & M, F, N & Voice, & \textit{ns} & “What would you describe the robot&\\
et al. (2020) &  &Name&  &  in the video as being?” Participants &\\
&  &  &  & could select either “a male robot”, “a &\\
&  &  &  &  female robot”, or “neither a male nor&\\
&  &  &  & a female robot”&\\[2ex]
Calvo-Barajas & M, F & Facial  & No & \textit{not applicable}&\\
et al. (2020) \cite{calvo2020effects} &&Features&&&\\[2ex]
Chita-Tegmark& M, F & Voice,  & No & \textit{not applicable}&\\
 et al. (2019) \cite{chita2019gender} &&Name&&&\\[2ex]
Chita-Tegmark& M, F & Name & No & \textit{not applicable}&\\
 et al. (2019) \cite{chita2019gender} &&&&&\\[2ex]
Eyssel \& Hegel & M, F & Facial & \textbf{Yes} & Participants rated the extent to which&\\
 (2012) \cite{eyssel2012s}& & Features,  & & the robot appeared “rather male” vs.&\\
&  & Hairstyle&  & “rather female” using a 7-point Likert&\\
&  & &  &  scale&\\[2ex]
Eyssel et al.& M, F & Voice & \textbf{Yes}*\textsuperscript{1}& Participants indicated whether the voice &\\
(2012a) \cite{eyssel2012if}& & & & sounded rather female (1) or male (7)&\\
& & & & using a 7-point Likert scale\\[2ex]
Eyssel et al. & M, F & Voice & \textbf{Yes} & Participants indicated whether the voice &\\
(2012b) \cite{eyssel2012activating}& & & & sounded rather female (1) or male (7)&\\
& & & & using a 7-point Likert scale&\\[2ex]
Ghazali et al. & M, F & Facial & \textbf{Yes} & Participants rated the robot's perceived&\\
(2018) \cite{ghazali2018effects} && Features&& gender on a semantic differential scale& \\
 &  &  Voice& & with end-points masculine/feminine &\\
 &  &  & & (supposedly 7-point Likert scale)&\\[2ex]
Jackson et al.& M, F & Voice,& No & \textit{not applicable}&\\
(2020) \cite{jackson2020exploring} &&Name&&&\\[2ex]
Jung et al. & M, F, & Clothes, & \textit{Yes} & Participants rated the robot's perceived &in the \textit{no cue},\\
(2016) \cite{jung2016feminizing} &\textit{no cue}&Color&& gender on a semantic differential item: & the robot was\\
&  & & &  “would you say that the robot was more & perceived as \\
&&&& like a male or like a female?” (1=male,&more male than\\
& & & & 7=female)& the male robot.\\[2ex]
Kraus et al. & M, F & Name, & \textbf{Yes} & (not disclosed) & the scale was\\
(2019) \cite{kraus2018effects} &&Voice&&&not disclosed.\\
&&&&&The paper only\\
&&&&&reports p-values\\
[2ex]
Kuchenbrandt & M, F & Name, & \textbf{Yes} & Participants indicated on a 7-point Likert &\\
et al. (2014) \cite{kuchenbrandt2014keep} & & Voice& &scale whether they perceived the robot as&\\
& & & &  being more female or male (1= more &\\
&&&&female; 7= more male)& \\[2ex]
Law et al. & M, F & Name, & No & \textit{not applicable}&\\
(2020) \cite{law2021interplay}&&Voice&&&\\[2ex]
Lugrin et al. & M, F & Voice & No & \textit{not applicable}&\\
(2020) \cite{lugrin2020if} &&&&& \\[2ex]
\multicolumn{6}{c}{\textit{(The table continues in the next page)}}\\[2ex]
\end{tabular}
\label{table:GenderManipulation1}
\end{table*}
\begin{table*}[]
\centering
\begin{tabular}{ l | l | l | l | l| l }
\multicolumn{6}{c}{\textit{(The table continues from previous page)}} \\[2ex]
\textbf{Authors} & \textbf{Robot's} & \textbf{Cues} & \textbf{Manip.} & \textbf{Metric} & \textbf{Notes} \\
 & \textbf{Genders} & \textbf{Used} & \textbf{Check} & \textbf{Used} &\\[2ex]
\hline\noalign{\smallskip}
Makenova & M, F & Name, & No & \textit{not applicable}&\\
et al. (2018) \cite{makenova2018exploring}&&Voice&&&\\[2ex]
Nomura \& & M, F & Name, & No & \textit{not applicable}&\\
Kinoshita (2015) \cite{nomura2015gender} &&Voice&&&\\[2ex]
Nomura \& & M, F & Name & \textbf{Yes} & Participants rated the adjectives &\\
Takagi (2011) \cite{nomura2011exploring} &&&& masculine and feminine on a&\\
&&&&  7-point Likert scale&\\[2ex]
Paetzel & M, F, N & Facial & \textit{Yes} & Participants rated the robot on a & no significant\\
et al. (2016a) \cite{paetzel2016congruency} &  & Features, & & 7-point Likert scale on the dimension&difference\\
&&Voice&& gender (supposedly 1=masculine,&between male \\
&&&& 7=feminine)&and neutral\\
&&&&&face\\[2ex]
Paetzel  & M, F & Facial & \textit{Yes} & Participants rated the perceived & the incongruent \\
et al. (2016b) \cite{paetzel2016effects} &  & Features, & & masculinity and femininity of&conditions  \\
&&Voice&&the robot on two Likert scales& differed in \\
&&&&&femininity but\\
&&&&&not in \\
&&&&&masculinity\\[2ex]
Pfeifer \& Lugrin & M, F & Name, & No & \textit{not applicable}&\\
(2018) \cite{pfeifer2018female}&&Voice&&&\\[2ex]
Powers \& Kiesler & M, F & Voice & \textbf{Yes} & Participants were asked whether &\\
(2006) \cite{powers2006advisor}&&&& the robot was male or female& \\
&&&& (for voice) and which name &\\
&&&& they would suggest for the &\\ &&&&robot&\\[2ex]
Powers et al.  & M, F & Facial& \textbf{Yes} & Participants filled out a write-in &\\
(2005) \cite{powers2005eliciting} &&Features,&&question: Is the robot gendered? &\\
&&Voice&& If yes, they were asked to rate & \\
&&&&how masculine and how feminine&\\
&&&& the robot was on a 5-point &\\
&&&& rating scale (1 = low, 5 = high) &\\[2ex]
Rea et al. (2015) \cite{rea2015check} & M, F & Pronouns & \textbf{Yes} & Participants rated the robot's gender & \\
&  &  & & on a 7-point Likert scale (feminine & \\
&  &  & & to masculine) & \\[2ex]
Reich-Stiebert \&  & M, F & Name, & \textbf{Yes} &Participants indicated the robot's&\\
Eyssel (2017) \cite{reich2017ir} &&Voice&&gender on a 7-point Likert scale &\\
&&&& (1=female, 7=male)&\\[2ex]
Sandygulova \cite{sandygulova2018age} & M, F & Voice & \textit{ns}& Participants rated the robot's & \\
\& O'Hare (2018)&&&& genderedness by choosing &\\
&&&&between three options: male,&\\
&&&&female, and not sure&\\[2ex]
Sandygulova& M, F & Name, & No & \textit{not applicable}&\\
\& O'Hare (2016) \cite{sandygulova2016investigating}&&Voice&&&\\[2ex]
Sandygulova & M, F & Voice & \textbf{Yes} & Children indicated the robot's&\\
\& O'Hare (2015) \cite{sandygulova2015children} &&&&genderedness (female and male)&\\
&&&&through a pictorial questionnaire&\\ [2ex]
Sandygulova  & M, F & Voice & No & \textit{not applicable}&\\
et al. (2014) \cite{sandygulova2014investigating}&&&&&\\[2ex]
Siegel et al.& M, F & Voice & No & \textit{not applicable}&\\
 (2009) \cite{siegel2009persuasive} &&&&&\\[2ex]
Tay et al. & M, F & Name, & \textbf{Yes}&Participants rated the robot's  & unclear how \\
(2014) \cite{tay2014stereotypes}&&Voice&&perceived masculinity and femininity &many points\\
&&&&on a Likert scale&compose the\\
&&&&&Likert scale\\[2ex]
\multicolumn{6}{c}{\textit{(The table continues in the next page)}}\\[2ex]
\end{tabular}
\label{table:GenderManipulation2}
\end{table*}
\begin{table*}[]
\centering
\begin{tabular}{ l | l | l | l | l| l }
\multicolumn{6}{c}{\textit{(The table continues from previous page)}} \\[2ex]
\textbf{Authors} & \textbf{Robot's} & \textbf{Cues} & \textbf{Manip.} & \textbf{Metric} & \textbf{Notes} \\
 & \textbf{Genders} & \textbf{Used} & \textbf{Check} & \textbf{Used} &\\[2ex]
\hline\noalign{\smallskip}
Thellman & M, F& Name, & Yes*\textsuperscript{2}&Participants rated the robot's genderedness&\\
et al. (2018) \cite{thellman2018he}& & Voice, && on a 7-point Likert scale spanning from&\\
& & Clothes, && 1 (female) to 7 (male)&\\
& &Color &&&\\[2ex]
You \& Lin & M, F, N & Voice, & No & \textit{not applicable}&\\
(2019) \cite{you2019gendered}& & Hairstyle,  & &&\\
& & Color & &&\\[2ex]
Zhumabekova  & M, F & Name, & \textit{ns} & Participants were asked what gender& details about the\\
et al. (2018) \cite{zhumabekova2018exploring} & & Voice, &&the robot was& measures used \\
& & Clothes &&& are missing\\[2ex]
Steinhaeusser & M, F, N & Voice & No & \textit{not applicable}&\\
et al. (2021) \cite{steinhaeusser2021anthropomorphize} & &&&&\\[2ex]
\hline
\multicolumn{6}{l}{*\textsuperscript{1}\textit{The statistical significance in \cite{eyssel2012if} is inferred from \cite{eyssel2012activating} which is based on the same study, but not directly reported.}}\\
\multicolumn{6}{l}{*\textsuperscript{2}\textit{Thellman et al. \cite{thellman2018he} do not report the results of the statistical analyses related to the manipulation check.}}
    \end{tabular}
    \label{Table:GenderManipulation3}
\end{table*}

\subsubsection{Robots}

In terms of robot choice, NAO was the most used robot (37\% of the papers, see Table \ref{table:General and demographic information}) followed by Furhat and Flobi (featured in 9\% of the papers each); Meka M1, Reeti, Willow Garage PR2, and Robovie (featured in 6\% papers each); and, finally, Alpha 1 Pro, Pepper, Socibot, and Nexi (featured in 3\% of the papers each). Four papers did not specify robotic platform used in the studies (11\% of the papers). In 65.7\% of the included papers, the robot was presented to participants through a physical embodiment, in 25.7\% of the studies through a video (although \cite{chita2019gender} use a video-recording of pictures), and in 8.6\% of the studies through images.

\subsection{Tasks and Activities}
\label{taskandactivities}

In this section, we report the tasks participants were asked to perform in the reviewed studies, as well as the activities the gendered robots were involved in. 

In \textit{static image studies} (cf. \textit{pictures} in Table \ref{table:General and demographic information}), participants were asked to carefully look at a picture of the robot and rate their perception of it on the relevant dependent variables \cite{bernotat2017shape,bernotat2021fe,eyssel2012s}. Similarly, in \textit{video-recording studies} (cf. \textit{video} in Table \ref{table:General and demographic information}), participants were asked to watch a short video of the robot and fill out a questionnaire. Some of the videos featured the robot speaking to the camera (e.g., explaining a topic) \cite{bryant2020should,eyssel2012if,eyssel2012activating,lugrin2020if,powers2006advisor,steinhaeusser2021anthropomorphize}. Others showed an actual interaction \cite{jackson2020exploring} or described it through a series of vignettes \cite{chita2019gender,law2021interplay}. In \textit{studies including a physical robot} (cf. \textit{physical} in Table \ref{table:General and demographic information}), participants observed a co-present physical robot performing a (set of) behavior(s) or explaining a topic \cite{calvo2020effects,makenova2018exploring,nomura2011exploring,nomura2015gender,paetzel2016congruency,paetzel2016effects,sandygulova2015children,siegel2009persuasive,thellman2018he,you2019gendered} or directly interacted with the robot \cite{ghazali2018effects,jung2016feminizing,kraus2018effects,kuchenbrandt2014keep,pfeifer2018female,powers2005eliciting,rea2015check,reich2017ir,sandygulova2014investigating,sandygulova2016investigating,sandygulova2018age,tay2014stereotypes,zhumabekova2018exploring}. They rated their perceptions of the robot and/or interaction immediately after. 

In the following, we briefly describe the content of the activities in the reviewed studies. In doing so, we focus only on those studies featuring a video-recorded or co-present demo or a video-recorded or first-person interaction and filter out those where the robot is used as a stimulus, for instance, to display an interactive behavior (e.g., facial expressions). We made this type of decision to be sure to present those interactions that had a more or less pronounced social context.

In the \textit{demo} studies, the robot introduced a topic to a co-present audience or an audience asynchronously watching. Siegel et al. \cite{siegel2009persuasive} used the robot to provide a brief explanation of its hardware, software, and technical abilities, and ask for donations. Makenova et al. \cite{makenova2018exploring} and You and Lin \cite{you2019gendered} replicated Siegel et al.'s study using the robot to introduce a research project and ask for donations \cite{makenova2018exploring} or to give an overview of the research taking place in the lab and ask for donations \cite{you2019gendered}.
Nomura and Kinoshita \cite{nomura2015gender} employed the robot to describe the construction of a commercial building, while Powers and Kiesler \cite{powers2006advisor} and Thellman et al. \cite{thellman2018he} used it to give health advice to participants \cite{powers2006advisor} or explain why humans should not be afraid of robots \cite{thellman2018he}. Finally, Sandygulova and O'Hare \cite{sandygulova2015children} and Steinhaeusser et al. \cite{steinhaeusser2021anthropomorphize} employed the robot to tell a story.

In presenting the \textit{interaction} studies, we first introduce the video-recorded studies, in which the interaction was only observed by the participants, and then the first-person interaction studies, in which the participants themselves took part in the interaction.
Three papers asked participants to \textit{observe} or read about an interaction: Chita-Tegmark et al. \cite{chita2019gender}, Jackson et al. \cite{jackson2020exploring}, and Law et al. \cite{law2021interplay}. All three papers included very complex interactions, which would have been difficult to carry through in a co-present human-robot interaction study.
Chita-Tegmark et al. \cite{chita2019gender} and Law et al. \cite{law2021interplay} used the exact same interaction in their studies and presented it through a series of vignettes in a video. The interaction takes place in an office setting between three characters: a supervisor and two subordinates.
In the interaction, the supervisor reproaches one of the subordinates for a mistake, and then leaves the room. The two subordinates, who are left in the room, discuss the situation and the subordinate who was not reproached (a human or a robot depending on the condition) reacts to the one who made the mistake in either a friendly or unfriendly way. Jackson et al. \cite{jackson2020exploring} presented the interaction through a video of a human-robot interaction. In the video, the robot explains how to play the game battleship and then supervises two humans while they play. At some point during the play, one of the humans receives a call and leaves the room. The human left in the room presents the robot with a morally problematic request, which the robot rejects in different ways.

Thirteen papers featured an actual \textit{first-person} interaction. In Ghazali et al. \cite{ghazali2018effects}, participants played a trust game inspired by the investment game, where they prepared a drink for an alien with the help of the robot.
In Jung et al. \cite{jung2016feminizing}, they interacted with the robot in a music listening scenario, whereas in Pfeifer and Lugrin \cite{pfeifer2018female}, they learned how to develop a website in HTML together with the robot.
In Kraus et al \cite{kraus2018effects}, Powers et al. \cite{powers2005eliciting}, and Rea et al. \cite{rea2015check} participants engaged in a conversation with the robot. In these studies, the robot acted as a dialogue partner in a taxi ordering or baby healthcare scenario \cite{kraus2018effects}, 
engaged participants in a face-to-face conversation on the topic of first dates \cite{powers2005eliciting}, or involved them in a casual conversation around daily topics (e.g., hobbies, work, or school) \cite{rea2015check}.  
Kuchenbrandt et al. \cite{kuchenbrandt2014keep} and Reich-Stiebert et al. \cite{reich2017ir} involved participants in more structured tasks. The former \cite{kuchenbrandt2014keep} asked participants to sort out items into the compartments of a sewing or tool box under the instruction of the robot. 
The latter \cite{reich2017ir} asked them to solve a set of cognitive tasks (i.e., a memory, an auditory, and a visual task) focusing on stereotypical female or stereotypical male academic fields.
Following the line of studies involving participants in stereotypical female or male tasks, Tay et al. \cite{tay2014stereotypes} engaged participants in either a healthcare scenario in which, among other things, the robot measured their body temperature, or in a safety scenario in which it, for instance, enlisted their help in resolving an intrusion in the research space.

The studies by Sandygulova et al. \cite{sandygulova2014investigating}, Sandygulova and O'Hare \cite{sandygulova2016investigating,sandygulova2018age}, and Zhumabekova et al. \cite{zhumabekova2018exploring} focused on interactions between children and robots. In \cite{sandygulova2014investigating}, the children were asked to help the robot practice its new job of keeping people safe by turning off kitchen appliances. In \cite{sandygulova2016investigating}, they were asked to help the robot learn how to use the utensils in the kitchen. In \cite{makenova2018exploring}, they were asked to help the robot lay the table. Finally, in \cite{sandygulova2018age}, the children interacted with the robot in three sessions. In the first two, they were involved in a card-pairing task. In the last one, they listened to the robot telling a story.

\subsection{Definition of and Motivation for Using Gender}
\label{definitions_motivations}

\begin{figure*}
    \centering
    \includegraphics[width=.8\textwidth]{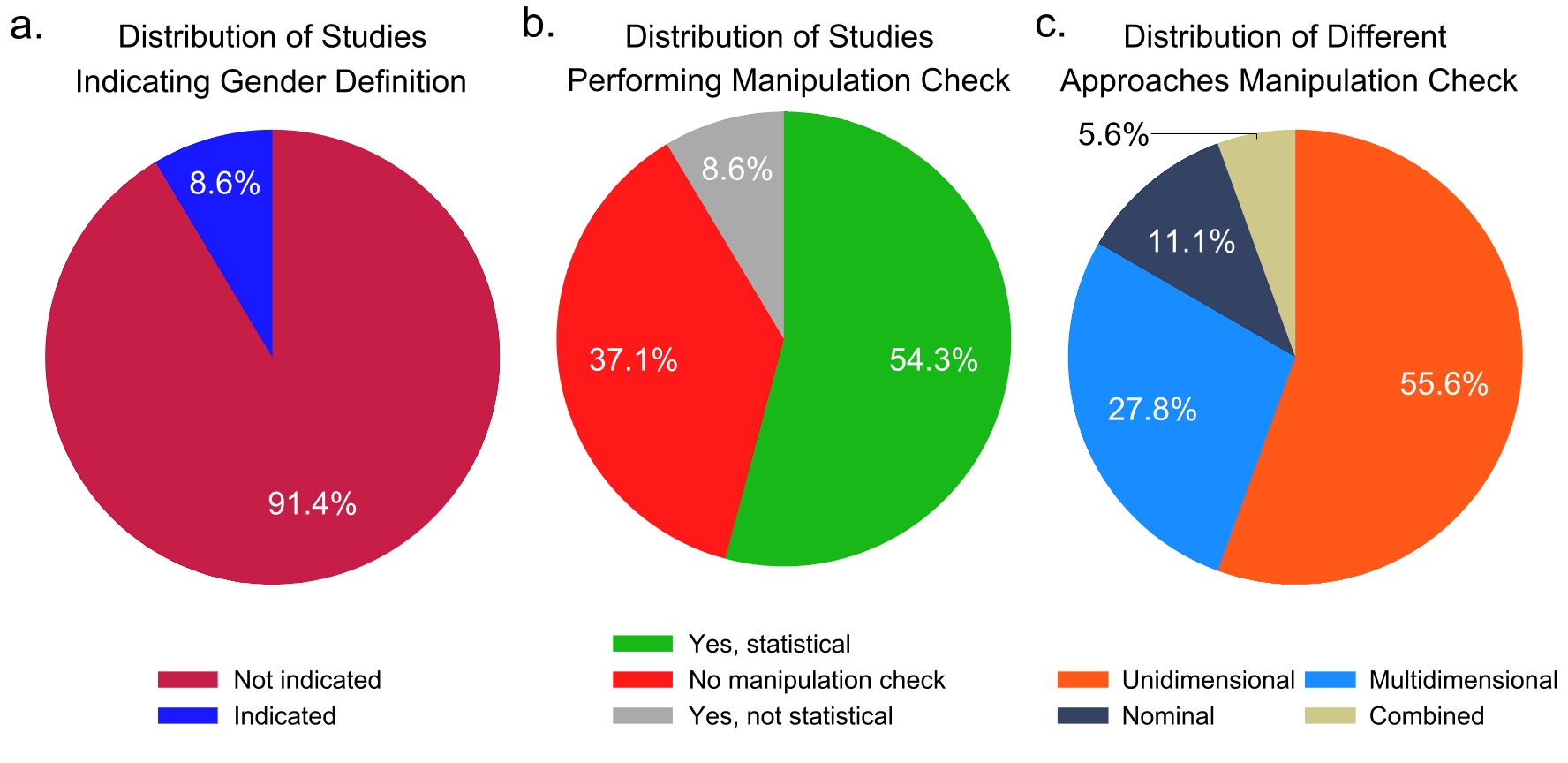}
    \caption{Percentage of studies indicating a gender definition (a), percentage of studies performing a manipulation check (b) and frequency of the different assessment approaches of the robot's ``gender" in the studies performing a manipulation check (c).}
    \label{fig:manipulation_gender}
\end{figure*}

\subsubsection{Definitions of Gender}

Most of the papers (91\%) did not provide a definition of gender or an explanation of the authors' understanding of gender (see Figure \ref{fig:manipulation_gender}a). One of them reported a definition of gender\textit{ing} \cite{bryant2020should}. Bryant et al. borrowed the term gendering from Robertson et al. \cite{robertson2010gendering} and defined it as ``the attribution of gender onto a robotic platform via voice, name, physique, or other features.'' They used this term to describe the \textit{encoding} of gender into robots via the choice of design features \cite{perugia2022shape} (see Section \ref{positionality}), rather than the property of the robot of being gendered.

Two other papers gave an explanation of their understanding of gender, both of them in relation to participants' gender. Rea et al. \cite{rea2015check} specified ``we use the term “gender” synonymously with biological sex, which we recognize is overly simplistic. We used "gender" for the practical purpose of simplifying our investigation." Reich-Stiebert and Eyssel \cite{reich2017ir}, instead, stated ``Sex refers to biological and physiological features. Gender, however, is a social construction." They explain that they included both of these factors in their experimental design as person’s biological sex might not correspond with their perceived gender identity. While these two definitions give us a clear understanding of the authors' interpretation of human gender, they do not provide us with their understanding of ``gender" or the process of gendering when it comes to robots.

\subsubsection{Reasons to Manipulate Robot Genderedness}
\label{reasonstomanipulate}

In terms of reasons to manipulate a robot's genderedness, we enlisted the rationale behind the robot's genderedness manipulation when explicitly mentioned by the authors. Jung et al \cite{jung2016feminizing}, Kraus et al. \cite{kraus2018effects}, Lugrin et al \cite{lugrin2020if}, Sandygulova \& O'Hare \cite{sandygulova2015children} Thellman et al. \cite{thellman2018he}, You and Lin \cite{you2019gendered}, and Zhumabekova et al. \cite{zhumabekova2018exploring} did not provide an explicit reason to manipulate the robot's genderedness. The other reviewed papers, instead, reported four core reasons behind the manipulation of the robot's genderedness. 

The first reported motivation was to study the relationship between social categorization and \textit{stereotypical judgements of robots}. In this group of papers, the robot's genderedness was manipulated to understand whether the robot's social categorization could elicit gender stereotypes \cite{bernotat2017shape,bernotat2021fe,eyssel2012s,nomura2015gender,rea2015check,reich2017ir}, bring people to attribute the robots capabilities in line with their perceived ``gender" \cite{bryant2020should,chita2019gender,kuchenbrandt2014keep,law2021interplay,powers2005eliciting,powers2006advisor}, or bring people to judge the appropriateness of the robots' behavior based on gender norms \cite{jackson2020exploring}.

The second reason was to study the influence of robot's genderedness on \textit{crucial HRI constructs}. In this group of papers, the robot's genderedness was manipulated to understand whether it could affect, among the others, people's acceptance of the robot \cite{eyssel2012if,eyssel2012activating,tay2014stereotypes}, their anxiety towards robots \cite{nomura2011exploring}, the robot's persuasiveness \cite{ghazali2018effects,makenova2018exploring,siegel2009persuasive}, trustworthiness, \cite{calvo2020effects,makenova2018exploring,siegel2009persuasive,tay2014stereotypes}, uncanniness \cite{paetzel2016congruency,paetzel2016effects}, and anthropomorphism \cite{eyssel2012if,eyssel2012activating,steinhaeusser2021anthropomorphize}. 

The third reason to manipulate the robot's genderedness was to investigate \textit{gender segregation} -- ``the separation of boys and girls into same-gender groups in their friendship and causal encounters'' \cite{mehta2009sex} -- in child-robot interaction (cHRI). In this group of papers, the robot's genderedness was manipulated to explore whether children retained gender segregation with gendered robots \cite{sandygulova2018age} and whether their preference for a same-gender robot changed across age and gender groups \cite{sandygulova2014investigating,sandygulova2016investigating}.
Finally, the fourth motivation was to test whether female social robots could be used as \textit{role models} to engage young women in computer science \cite{pfeifer2018female} Since Denner et al. \cite{denner2005girls} showed that girls benefit from learning how to program in female pairs, Pfeifer and Lugrin wanted to understand whether the genderedness of the robot could impact the learning process of women in the domain of computer science. 

\subsection{Gender Manipulation (RQ1)}
\label{gendermanipulation}

\subsubsection{Voice} 
\begin{figure*}
   \centering
    \includegraphics[width=.85\textwidth]{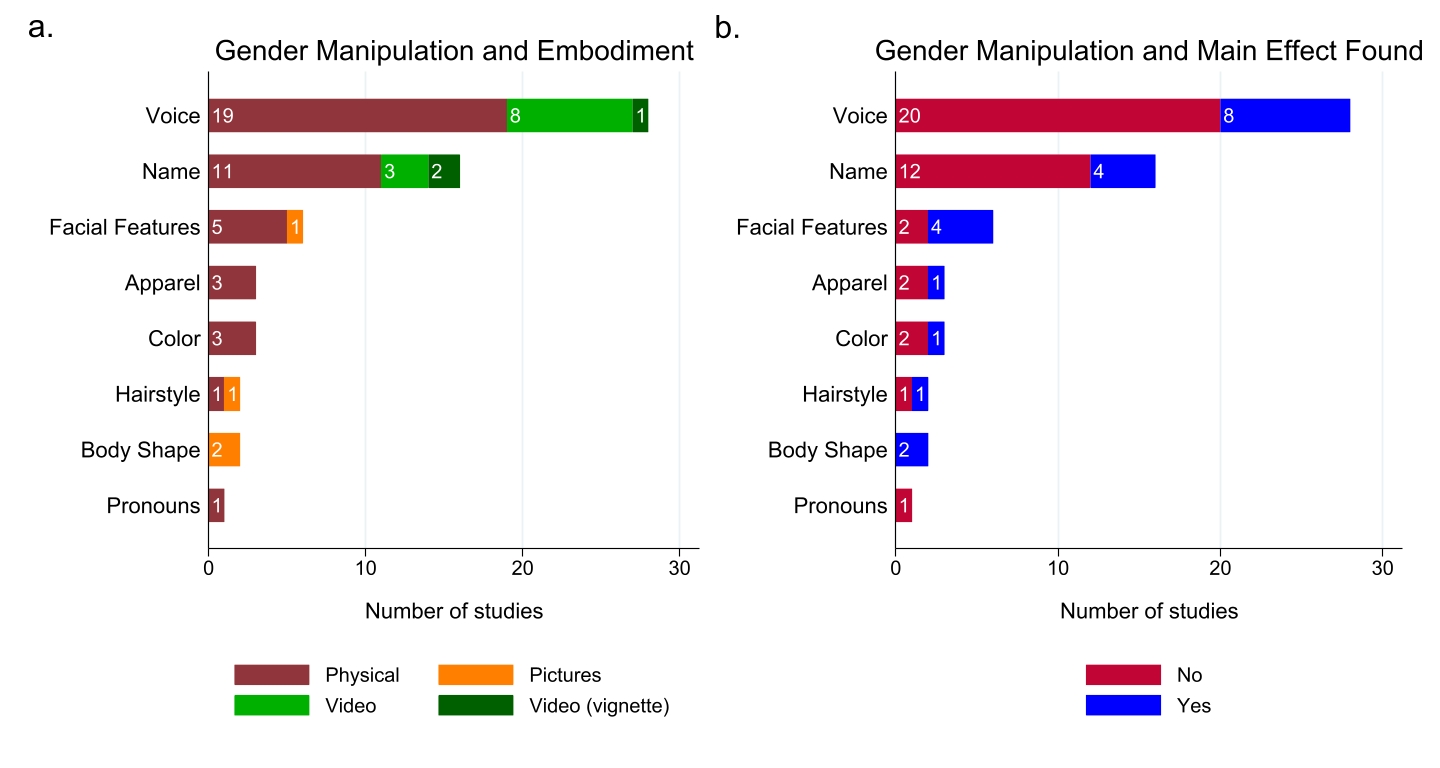}
    \caption{\textbf{Frequency of Manipulations.} 4a. Different 
    manipulations in decreasing order of frequency and type of embodiment used in the studies, 4b. different manipulations in decreasing order of frequency and corresponding significant (or not) main effect of robot's genderedness on the dependent variables.}
    \label{fig:embodiment}  
\end{figure*}

In terms of design choices, 28 studies (78\%, see Table \ref{table:GenderManipulation1} and Figure \ref{fig:embodiment}) manipulated the robot's genderedness through its voice, either in isolation ($N=9$) or in combination with other features ($N=19$, we report the combinations in the other sections). In most cases, the voices used were the default female and male voices provided by commercially available text-to-speech software, such as MacOS' \cite{law2021interplay}, CereProc \cite{paetzel2016effects}, Cepstral Theta \cite{powers2006advisor}, Acapella \cite{thellman2018he}, or voices edited with software like Audacity \cite{steinhaeusser2021anthropomorphize}. In other cases, human voices were recorded and implemented on a robot (e.g., Sandygulova et al. \cite{lugrin2020if}). 

Since the voices employed in the reviewed studies were in most cases the default voices provided by commercially available software, the majority of authors did not specify the rationale behind their selection. Only Kuchenbrandt et al. \cite{kuchenbrandt2014keep} mentioned low frequency as the main characteristic of male voices and high frequency as the characterizing feature of female voices, and Powers and Kiesler \cite{powers2006advisor} and Sandygulova and O'Hare \cite{sandygulova2015children} mentioned work by Nass and Brave \cite{nass2005wired} explaining how a voice with a fundamental frequency of $\approx$110 Hz is perceived as male and a voice with a fundamental frequency of $\approx$210 Hz as female.

\subsubsection{Name \& Pronouns}
Sixteen studies (44\%) employed gendered names to manipulate the robot's genderedness. Names were used in isolation ($N=2$ \cite{chita2019gender,nomura2011exploring}), in combination with voice alone ($N=12$\cite{bryant2020should,chita2019gender,jackson2020exploring,kraus2018effects,kuchenbrandt2014keep,law2021interplay,makenova2018exploring,nomura2015gender,pfeifer2018female,reich2017ir,sandygulova2016investigating,tay2014stereotypes}), or in combination with voice and other features ($N=2$; voice and clothes \cite{zhumabekova2018exploring}; voice, clothes, and color \cite{thellman2018he}). The rationale to use names to manipulate a robot's genderedness is never explained in detail in the studies we reviewed. Among the names used, we found James and Mary \cite{bryant2020should}, Bob and Alice \cite{jackson2020exploring}, Nero and Nera \cite{kuchenbrandt2014keep}, Peter and Katie \cite{law2021interplay}, Robie/Ruslan and Rosie/Roza \cite{makenova2018exploring}, Taro and Hanako \cite{nomura2015gender}, Lena and Leon \cite{pfeifer2018female}, Robie and Rosie \cite{sandygulova2016investigating,zhumabekova2018exploring}, and John and Joan \cite{tay2014stereotypes}. Rea et al. \cite{rea2015check} used the gender neutral name Taylor for both robot's ``genders" and manipulated genderedness with the pronouns she/he. They were the only ones manipulating genderedness this way (See Figure \ref{fig:embodiment}).

\subsubsection{Facial Features} \label{facialfeatures}
Six studies (17\%) employed facial features to manipulate the robot's gender. Within this category, there was a lot of variability in terms of what facial elements were used to manipulate the robot's genderedness. For instance, Eyssel and Hegel \cite{eyssel2012s} used Flobi's lip module with more defined lips to manipulate the genderedness of the female robot, and the one with less defined lips to manipulate the genderedness of the male robot. Powers et al., \cite{powers2005eliciting} instead, used the color of the lips to change the perception of the robot's genderedness: pink lips for the female robot and grey lips for the male one.

At a more holistic level, Calvo-Barajas et al. \cite{calvo2020effects} and Ghazali et al. \cite{ghazali2018effects} used the default faces provided by the robots Furhat and Socibot. In both their studies, the female texture had thinner eyebrows, rosier cheeks, and redder lips than the male texture. Paetzel et al. \cite{paetzel2016congruency,paetzel2016effects} did not resort to Furhat's predefined faces. They used the software FaceGen to create the female and male facial textures they then projected onto Furhat's face mask. The software FaceGen gives the possibility to model a 3D head and modify its genderedness through a slider. From the pictures shared by the authors, it seems that the female texture had thinner eyebrows, redder lips, bigger eyes, and a whiter skin with respect to the male texture, all facial features partly overlapping with those in Calvo-Barajas et al. and Ghazali et al. 

Facial features appear in isolation only once and are combined with the robot's hairstyle in Eyssel and Hegel \cite{eyssel2012s} and with the robot's voice in 4 studies \cite{ghazali2018effects,paetzel2016congruency,paetzel2016effects,powers2005eliciting}. Interestingly, the choice of facial features used to manipulate the robot's genderedness is never explained in detail or motivated by the studies. This might have to do with the fact that in most studies the faces used to manipulate the robot's genderedness were the default faces provided by the respective robotic platforms (i.e., Furhat and Socibot). Hence, the authors of the papers might have worked under the assumption that a rationale for the choice of facial features had been followed by the respective robotic companies.

\subsubsection{Apparel \& Color}
\label{apparelcolor}

Three studies (8\%) used clothes to manipulate the robot's genderedness. Jung et al. \cite{jung2016feminizing} provided the male robot with a man's hat and the female robot with pink earmuffs. Thellman et al. \cite{thellman2018he} equipped the male robot with a blue white-dotted bow tie and the female robot with a pink ribbon. Finally, Zhumabekova et al. \cite{zhumabekova2018exploring} gave the female robot a flower hair clip and the male robot a bow-tie. Clothes were used in combination with voice and names in \cite{thellman2018he,zhumabekova2018exploring}. Jung et al. did not give details regarding other gender cues beyond clothes. However, we suspect that they also used the robot's voice to manipulate the robot's genderedness as the robot had a conversation with participants in their scenario. 

The clothes in the reviewed studies were often stereotypically colored (color is used in 3 studies, 8\%): blue for male robots, pink for female robots \cite{jung2016feminizing,thellman2018he}. In You and Lin \cite{you2019gendered}, it is the body of the robot that is stereotypically colored instead: blue for the male robot, grey for the neutral robot, and pink for the female robot. The rationale behind using clothes and color to manipulate robot's genderedness is never explicitly laid down.

\subsubsection{Hairstyle} 

Two studies (6\%) employed the robot's hairstyle to suggest the robot's genderedness. Eyssel and Hegel \cite{eyssel2012s} used Flobi's hair module to add short or long hair to the robot, whereas You and Lin \cite{you2019gendered} used the robot Alpha 1 pro with short, mid-length, and long hair to manipulate female, neutral, and male genderedness respectively. 
While You and Lin did not provide any rationale for their manipulation of genderedness, Eyssel and Hegel mentioned Brown and Perrett \cite{brown1993gives}, and Burton et al. \cite{burton1993s} to justify the choice of using hair length. These papers pose that hairstyle is a salient facial cue to determine someone's gender and that long hair lead to an increased accessibility of knowledge structures about the social category of women, whereas short hair activate stereotypical knowledge structures about men. In Eyssel and Hegel \cite{eyssel2012s}, the robot's hairstyle is used in combination with its facial features (see Section \ref{facialfeatures}), while in You and Lin \cite{you2019gendered} with the robot's voice and color (see Section \ref{apparelcolor}).

\subsubsection{Body Shape}

Two studies (6\%) used the robot's body proportions to manipulate the robot's genderedness. These studies were both authored by Bernotat et al. \cite{bernotat2017shape,bernotat2021fe} and the latest of the two was a replication of the earliest. Bernotat et al. modified the Waist-to-Hips Ratio (WHR) and Shoulder Width (SW) of a robot's drawing to achieve different perceptions of genderedness. They hypothesized that a robot with a WHR of 0.9 and a SW of 100\% would be perceived as male, whereas a robot with a WHR of 0.5 and 80\% SW as female. The rationale behind this manipulation of genderedness came from the work of Johnson and Tassinary \cite{johnson2005perceiving} and Lippa \cite{lippa1983sex} who showed that people rely on WHR to judge a target’s ``gender" and that the form of the waist is a relevant feature for gender perception. Since the studies used static images, body proportions were not used in combination with other cues.

\subsection{Manipulation Check \& Assessment Tools}
\label{manipulationANDassessment}

Only 54.3\% of the studies ($N=19$) performed \textit{statistical} analyses to understand whether the manipulation of the robot's genderedness actually succeeded (see Figures \ref{fig:manipulation_gender}b and \ref{fig:ResultsReview}). On top of these studies, 8.6\% of the studies ($N=3$) performed a manipulation check but of a \textit{non-statistical} nature \cite{bryant2020should,sandygulova2018age,zhumabekova2018exploring} (see Figures \ref{fig:manipulation_gender}b and \ref{fig:ResultsReview}). The authors did ask participants which ``gender" the robot belonged to in their opinion, but they did not perform any statistical analysis to check for the significance of the result. As is easy to infer, 37.1\% of the reviewed studies ($N=13$) did not perform any manipulation check to test whether participants perceived the robot's genderedness as expected \cite{calvo2020effects,chita2019gender,jackson2020exploring,law2021interplay,makenova2018exploring,nomura2015gender,pfeifer2018female,sandygulova2016investigating,sandygulova2015children,sandygulova2014investigating,siegel2009persuasive,you2019gendered,steinhaeusser2021anthropomorphize}. 

In the studies that performed a statistical manipulation check, the authors used three different approaches to assess people's attribution of ``gender" to the robot (See Table \ref{table:GenderManipulation1} and Figure \ref{fig:manipulation_gender}c). 
The first measurement approach was \textit{unidimensional}. The authors asked participants to rate the robot's genderedness on one item usually using the following phrase: \textit{Rate the extent to which the robot appeared “rather male” versus “rather female”}. The rating was expressed on a 7-point Likert scale with \textit{male} and \textit{female} as end points.  
The second measurement approach was \textit{multidimensional} (See Table \ref{table:GenderManipulation1} and Figure \ref{fig:manipulation_gender}c). The authors asked participants to fill out two items usually using the following phrasing: (1) \textit{To what extent do you perceive the robot as male?} (2) \textit{To what extent do you perceive the robot as female?}. The ratings were expressed on 7-point Likert scales where 1 meant \textit{not at all} and 7 \textit{extremely} \cite{bernotat2017shape,bernotat2021fe,nomura2011exploring,paetzel2016effects,powers2005eliciting,tay2014stereotypes}. 
Finally, the third and last measurement approach was \textit{nominal} (See Table \ref{table:GenderManipulation1} and Figure \ref{fig:manipulation_gender}c). The authors asked participants to select the ``gender" of the robot among a list of options or as a write-in question \cite{powers2005eliciting,powers2006advisor,sandygulova2015children}. Sandygulova and O'Hare used this approach with children using a pictorial response system \cite{sandygulova2015children}. Powers and Kielser \cite{powers2006advisor} asked participants to attribute a name to the robot and judged the ``gender" attributed to the robot based on the gender of the name. Finally, Powers et al. \cite{powers2005eliciting} combined the multidimensional and nominal approaches by first asking whether the robot in their study was gendered and then asking participants to specify how feminine and masculine the gendered robot was.

When Likert scales were used to measure the robot's genderedness (first and second approach), the mean scores on the items female/feminine and male/masculine were only rarely close to the end points of the corresponding gender. As an example, for Ghazali et al. \cite{ghazali2018effects}, the manipulation check was significant. However, the difference between the male and female robot was not marked (male robot: $M=5.50$, $SD=1.60$; female robot $M= 6.07$, $SD= 0.83$).  
When the manipulation of the robot's genderedness was performed with nominal scales (third approach), the difference between the robot's ``genders" was obviously more marked. However, female robots were more difficult to categorize across studies. This was particularly evident in \cite{powers2005eliciting} where the robot with the dampened female voice was miscategorized by 73\% of the participants and given a male name by 70\% of them. 

Overall, 79\% of the studies performing a statistical manipulation check ($N=15$) were successful in manipulating the robot's genderedness. Sixteen percent of them ($N=4$) were only partially successful. Finally, 5\% of them ($N=1$) did not report the results of the statistical manipulation check \cite{thellman2018he} (see Table \ref{table:GenderManipulation1} and Figure \ref{fig:ResultsReview}). The only instances where the manipulation check was only partially successful were the studies with a gender neutral or gender incongruent condition \cite{jung2016feminizing,paetzel2016congruency,paetzel2016effects}, or an altered gendered voice \cite{powers2006advisor}. 

\subsection{Results: Effects of Robot's Genderedness (RQ2)}
\label{effects}

\subsubsection{Methodological Note} 
The studies we reviewed employed 132 dependent variables. These could be nested into 17 groups based on conceptual similarity (e.g., warmth and mildness were nested under communion).
For convenience, we refer to the group variables when reporting main and interaction effects. This grouping was merely done to clearly summarize the results and draw conclusions from them.

\onecolumn
\begin{landscape}
\begin{center}
\begin{figure*}
\includegraphics[width=1.3\textwidth]{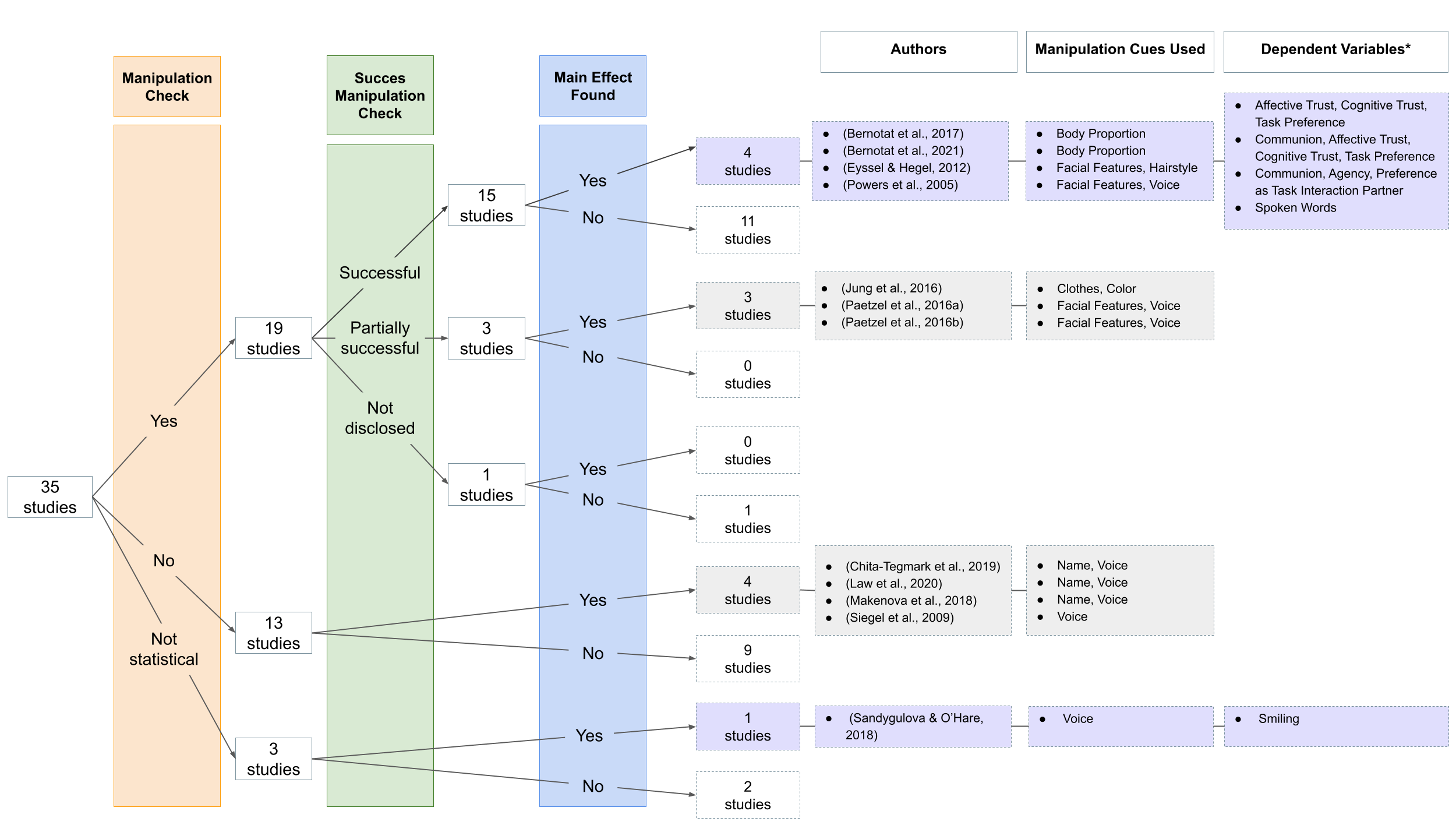}
    \caption{\textbf{Diagram Summarizing the Results of the Scoping Review.} The \textit{orange column} displays which of the included studies enlists a manipulation check, the \textit{green column} shows how many of the studies performing a manipulation check actually succeeded in manipulating the robot's genderedness, and the \textit{blue column} highlights the studies finding a main effect of the robot's genderedness on the dependent variables. The \textit{purple boxes }on the right enlist the papers featuring main effect of gender on the dependent variables, the gender cues used when such effect was found, and the dependent variables influenced by robot’s genderedness. *= the dependent variables reported here are only those significantly affected by the robot's genderedness.
}
    \label{fig:ResultsReview}
\end{figure*}
\end{center}
\end{landscape}
\twocolumn

\subsubsection{Main Effects}
\label{maineffects}

In the reviewed studies, only 17\% of the dependent variables (22 dependent variables out of 132) were affected by the manipulation of the robot's genderedness in terms of main effects. The genderedness of the robot did not yield any significant effect on the dependent variables nested under \textit{competence} (10 dependent variables), \textit{likability} (15 dependent variables), \textit{credibility} (3 
dependent variables), \textit{acceptance} (8 dependent variables), \textit{task-related robot evaluations} (4 dependent variables), \textit{proximity} (1 dependent variable), \textit{closeness} (2 dependent variables), and ``other" (2 dependent variables). Moreover, it had seldom main effects also on the dependent variables in the other groups. 

When the results were significant, participants tended to perceive the robot in line with \textit{gender stereotypes} (see Section \ref{reasonstomanipulate}). They attributed more communal traits to female robots than to male robots \cite{bernotat2021fe,eyssel2012s} (\cite{bernotat2017shape} marginally significant) and more agentic traits to male robots than to female robots \cite{eyssel2012s}. They showed higher affective trust towards female robots than towards male robots \cite{bernotat2017shape,bernotat2021fe}, and rated the female robot as more suitable for stereotypical ``female" tasks \cite{bernotat2017shape,bernotat2021fe,eyssel2012s} and the male robot as more suitable for stereotypically ``male” tasks  \cite{eyssel2012s}. Moreover, they donated more money \cite{makenova2018exploring,siegel2009persuasive}, said more words \cite{powers2005eliciting}, and smiled more to female robots than to male robots \cite{sandygulova2018age}. The only studies that were counterintuitive in terms of gender stereotypes were Chita-Tegmark et al.'s \cite{chita2019gender} where, in contrast with the authors' expectations, the male robot was perceived as more emotionally intelligent than the female one, and Bernotat et al.'s \cite{bernotat2017shape,bernotat2021fe}, where, as opposed to the author's assumptions, the female robot elicited more cognitive trust than the male robot.

Very few studies disclosed a significant main effect of the robot's genderedness on \textit{crucial HRI constructs} (see Section \ref{reasonstomanipulate}). In \cite{jung2016feminizing}, the female robot was rated significantly higher in animacy and anxiety than the male one, and in \cite{law2021interplay}, it was trusted significantly less. Interestingly, some of these studies report conflicting evidence. For instance, the male robot was perceived as more anthropomorphic than the female robot in \cite{jung2016feminizing}, while it was perceived as more machinelike in \cite{paetzel2016congruency}.

\subsubsection{Interaction Effects}
\label{interactioneffects}

The reviewed studies showed a significant interaction effect of the robot's genderedness and (an)other independent variable(s) on 24.24\% of the dependent variables (32 of the 132 dependent variables). Fifty percent of these effects resulted  from the interaction between the robot's genderedness and participant's gender. The other half of these effects resulted from the interaction between the robot's genderedness and a further independent variable (i.e., severity of moral infraction \cite{jackson2020exploring}, interaction modality \cite{paetzel2016congruency}, type of emotion \cite{calvo2020effects}, childlikeness of the robot \cite{powers2006advisor}, stereotypically gendered task \cite{kuchenbrandt2014keep}, or learning material \cite{pfeifer2018female}).

\paragraph{Robot's Genderedness and Participant's Gender.}
Among the studies that found an interaction effect between the robot's genderedness and the participants' gender, 50\% (8 out of 16 dependent variables) showed a significantly positive effect of the \textit{matching} between the robot's genderedness and the participant's gender, and 50\% (8 out of 16) the opposite, a significantly positive effect of the \textit{mismatch} between the robot's genderedness and the participant's gender.
With regards to the former results, adults seemed to perceive a robot with the same gender as them as significantly less harsh \cite{jackson2020exploring}, more anthropomorphic \cite{eyssel2012if}, more psychologically close \cite{eyssel2012if}, and eliciting less negative cognition \cite{ghazali2018effects}. Further results disclosed that children were in a significantly better mood \cite{sandygulova2018age}, smiled more \cite{sandygulova2014investigating}, played more \cite{sandygulova2016investigating}, and got more physically close \cite{sandygulova2016investigating} to a robot that shared the same gender as them, which lends support to the \textit{gender segregation hypothesis} for cHRI. No evidence was found in support of the use of female robots as \textit{role models} for women learning computer science topics \cite{pfeifer2018female}.

With regards to the positive effect of a human-robot gender mismatch, women seemed to attribute higher emotional intelligence to male robots \cite{chita2019gender} and men found female robots more trustworthy \cite{siegel2009persuasive}, credible \cite{you2019gendered} (although \cite{siegel2009persuasive} find this effect for both men and women), and engaging \cite{siegel2009persuasive} and were willing to donate them more money \cite{siegel2009persuasive}. Furthermore, men and women uttered more words to the robot of the opposite ``gender" in \cite{powers2005eliciting}, and younger children showed more happiness in the opposite gender than in the same gender condition in \cite{sandygulova2018age}. This latter is the only result that disconfirms the gender segregation hypothesis for cHRI. In general, the results of the studies exploring human-robot gender (mis)match on the perception and interaction with robots are inconclusive when it comes to adult participants.

\paragraph{Robot's Genderedness and Further Independent Variables.} Fifty percent of significant interaction effects were due to the joint effect of the robot's genderedness and another independent variable. In \cite{paetzel2016congruency}, the female robot was perceived as more responsible, intelligent, pleasant, relaxed, and content than the male robot, but only in the multimodal condition (i.e., when the robot used both facial expressions and voice to interact), whereas the male robot was perceived as more familiar and trustworthy than the female robot, but only in the unimodal condition (when it used only facial expressions). In \cite{calvo2020effects}, the male robot was perceived as more likable in terms of appearance when it expressed high anger (as opposed to medium anger) and low happiness (instead of medium anger, and low anger), while the female robot was perceived as less likable in terms of appearance when it expressed high anger (instead of all other emotions: low, medium, and high happiness, and low and medium anger) and low anger and medium happiness (instead of low happiness). In \cite{powers2006advisor}, 100\% of the participants said they would be willing to follow the advice of the childlike male robot, 91\% of the participants said they would be willing to follow the advice of the adultlike male and childlike female robots, and only 50\% of the participants said they would be willing to follow the advice of the adultlike female robot. In \cite{jackson2020exploring}, participants perceived the male robot as too direct in the pre-test but not when responding to norm violating commands, but did not perceive such a difference for the female robot. Moreover, male participants liked male robots when rejecting commands from male humans for severe norm violations, but did not like female robots rejecting commands from female humans for weak norm violations. Also, male participants liked male robots but not female robots when they issued strong rejections. Finally, female participants preferred when robots did not comply with the requests of a human with the same gender as the robot. In \cite{reich2017ir}, participants who were instructed to solve a stereotypically female task with a male robot and those who were instructed to solve a stereotypically male task with a female robot reported higher contact intentions with respect to participants involved in conditions where the genderedness of the task and the genderedness of the robot matched each other.

\section{Addendum: Papers 2021-2022}
To conclude our Results section, we would like to report a short addendum on the studies manipulating the robot's genderedness between May 2021 and May 2022. To identify the studies in this addendum, we used the same search strings and databases detailed in Section \ref{datacollection} and followed the same selection pipeline discussed in Section \ref{selection_pipeline}. However, we did not perform the full process of coding and information extraction described in Section \ref{coding}. The present section only aims at indicating the most recent developments in the investigation of robots' genderedness and highlighting whether novel results have been disclosed. The short review we performed returned 40 papers, of which 7 met the inclusion criteria after reading the abstract, and only 5 after reading the entire article \cite{forgas2022effects,neuteboom2021people,perugia2021gender,pitardi2022effects,seo2022female}. In Table \ref{tab:Addendum}, we give more details about these papers.

Neuteboom and de Graaf (2021) \cite{neuteboom2021people} looked into the effects of robot's genderedness (female and male robot) and task (analytical and social) on the robot's perceived trustworthiness (i.e., capacity trust and moral trust), as well as on its social perception (i.e., agency and communion), and humanness (i.e., human uniqueness and human nature). 
In line with previous studies, they did not find any significant effect of robot's genderedness and performed task on people's perceptions. 

Perugia et al (2021) \cite{perugia2021gender}, instead, explored how people attribute gender (femininity and masculinity) and stereotypical traits (communion and agency) to Furhat. 
Most Furhat's faces were attributed a ``gender" in line with their names. 
Interestingly, the robot's genderedness influenced people's perceptions of the robot's agency but not of its communion. This study confirms that the robot's genderedness can influence the attribution of stereotypical traits to humanoid robots in agreement with \cite{bernotat2017shape,bernotat2021fe,eyssel2012s}.

\begin{table}[b!]
\caption{Details about the studies in the addendum: Authors, cues used to manipulate the robot's genderedness, and dependent variables (in \textbf{bold}, the significant main effects).}
    \centering
    \begin{tabular}{r|l|l}
    \textbf{Authors} & \textbf{Cues Used} & \textbf{Main Effect}\\[2ex]
     \hline
       Neuteboom &  Name, & trustworthiness, social \\
      
      \& de Graaf \cite{neuteboom2021people} &  Apparel & perception, humanness\\[2ex]
    Perugia  & Voice, Fac.  & \textbf{agency}, communion\\ 
    et al. \cite{perugia2021gender} & features &  \\ [2ex]
    Forgas-Coll  & Voice, Fac. & intention to use, use-\\
    et al. \cite{forgas2022effects} & features &  fulness, ease of use,\\
    & & enjoyment, social influ-\\
    & & ence, adaptiveness,\\
    && sociability\\ [2ex]
    Pitardi et & Personal & cultural values, percei-\\
  al. \cite{pitardi2022effects}  & titles, color & ved control, feelings of \\
    & & comfort, service brand \\
    && attitude, familiarity \\[2ex]
   Seo et al. \cite{seo2022female} & Name & \textbf{pleasure}, \textbf{satisfaction}\\[2ex]
   \hline
    \end{tabular}
    \label{tab:Addendum}
\end{table}

The other three studies focused on the genderedness of service robots. 
Forgas-Coll et al. (2022) \cite{forgas2022effects} investigated the effects of gender-personality congruity on customers' intention to use a service robot. 
They discovered that while the congruous gender-personality robots (female-cooperative and male-competitive) did not differ from the incongruous ones (female-competitive and male-cooperative) in promoting intention to use, they did differ between each other: the female-cooperative robot performing significantly better than the male-competitive one in promoting intention to use. 

With a slightly similar objective, Pitardi et al. (2022) \cite{pitardi2022effects} looked into the effects of matching robot’s genderedness and participant’s gender on people's perceived comfort and control in a service encounter, as well as on their brand attitude (i.e., positive and negative evaluations of the service provider). 
The study disclosed that human-robot gender congruity has a significant positive influence on perceived control and comfort, but not on brand attitude, and that the cultural value of masculinity mediates the effect of human-robot gender congruity on participant's perception of control.

Again in a service context, Seo (2022) \cite{seo2022female} investigated the effects of robot's genderedness on pleasure and customer satisfaction in a service encounter and took into account the robot's anthropomorphism as an additional independent variable. 
The results showed that a female service robot leads to higher satisfaction and pleasure than a male service robot and that the robot's anthropomorphism plays a key role in positively influencing the results.

To sum up, the five studies in the addendum did not introduce novel ways of manipulating the genderedness of humanoid robots (except from personal titles, which can be equated to pronouns, see Table \ref{tab:Addendum}). In terms of results, however, they do disclose some interesting insights. 
They show a preference for female robots and human-robot gender congruity in service contexts \cite{forgas2022effects,pitardi2022effects,seo2022female}. Interestingly, they also reveal that values of masculinity play a role in this preference. It might be that service contexts are much more powerful than others in eliciting stereotypical knowledge of male and female roles, and especially so for those participants with more conservative views of gender.

\section{Discussion}
\label{discussion}

In the following, we are going to summarize the main findings of the literature review, answer the research questions, and identify gaps in the literature that warrant further attention. Then, we discuss the results of the review and provide guidelines that the HRI community could follow when \textit{gendering} or studying the gendering of robots. In doing so, we combine our epistemological backgrounds in Social Robotics and Gender Studies.

\subsection{Summary of Results \& Answers to RQ1 and RQ2}

To summarize the results of the scoping review, \textbf{the HRI scholarship most often manipulated the robot's genderedness through its voice, name, and facial features} (\textbf{RQ1}). These cues were mostly used in interactive studies enlisting the use of a physical robot (see Figure \ref{fig:embodiment}). In the majority of cases the manipulation of the robot's genderedness with voice, name, and facial features yielded the expected results in terms of gendered perceptions (i.e., successful manipulation check). However, it often failed to produce a main effect of the robot genderedness on the dependent variables. Indeed, if we take a look at Figure \ref{fig:embodiment}b and the purple boxes in Figure \ref{fig:ResultsReview}, we realize that the most successful gender cues in influencing people's perceptions of robots were body proportions \cite{bernotat2017shape,bernotat2021fe}, and facial features \cite{eyssel2012s,powers2005eliciting}. If we pay close attention to the results of this scoping review, what becomes apparent is that the studies enlisting a significant main effect of the robot's genderedness on the dependent variables are predominantly picture-based (e.g., communion, agency, task preference). Moreover, we can see that, in these studies, \textbf{robot's genderedness is mostly successful in eliciting gender stereotypes of communion, agency and task preference/suitability, but does not yield notable significant effects on crucial HRI constructs, such as competence, likability, and acceptance} (\textbf{RQ2}).

Given that robot's genderedness seems to be more harmful than useful as a design feature (it affects stereotyping but does not improve HRI), robotic companies might want to resort to less humanlike robots when gender stereotypical tasks are involved, or, in case humanlike robots cannot be avoided, they might want to use gender cues less prone to elicit gender stereotypes.   
\textit{Perugia et al. \cite{perugia2023models} started investigating which design cues in a robot are more likely to elicit stereotyping. However, more research in this direction is needed} (\textbf{GAP 1}). Besides, given stereotypes towards gendered robots are so prevalent but mostly studied with static images and in short-term studies, \textit{future HRI research should investigate if stereotype attribution is influenced by a robot's embodiment} (\textbf{GAP 2}) \textit{and whether it changes over time} (\textbf{GAP 3}). In a repeated interaction study, Paetzel et al. \cite{paetzel2020persistence} discovered that participants develop stable perceptions of a robot's warmth and competence (concepts similar to communion and agency) after two minutes of interaction and do not update them over time. Longitudinal perceptual studies like Paetzel et al.'s are needed also in the context of gendered HRI, to disclose whether stereotypes are formed once and for all a few minutes after meeting a robot or can modify with repeated interactions. In addition, since many studies focused on explicit stereotyping 
\textit{it might be worth performing implicit bias studies \cite{nosek2011implicit} investigating people's automatic, pre-reflective stereotyping of gendered robots} (\textbf{GAP 4}). Finally, since the main concern of Roboethics and Robophilosophy is that people's behaviors towards robots might eventually generalize to humans, \textit{the HRI scholarship is in need of research paradigms and studies that explore whether and how the gender stereotyping people display towards robots can influence their attitudes towards humans} (\textbf{GAP 5}). 

\subsection{Discussion of Methodological Pitfalls}

None of the studies we reviewed included non-binary, transgender, gender non-conforming, and gender fluid participants. Thirty-nine out of 3902 participants taking part in the reviewed studies (i.e., 1\%) selected the option other/undisclosed. We can only assume that part of these participants identified with a gender falling outside of the binary. We consider the lack of gender-diverse participants a huge gap when studying the process of gendering robots, especially considering that the studies in this review brought to light the complex interweavings of participants' gender and robot's genderedness.
This might have happened because participants' gender is oftentimes asked with check-boxes providing only two options, ``female'' and ``male'', 
but it might have also happened due to the lack of a proactive effort in including more gender identities. We advocate for this effort, hence we propose a first guideline for research on gendering robots:
\vspace{0.1cm}
\begin{quote}
\noindent\textbf{Guideline 1}: \textit{Include transgender, gender fluid, gender non-conforming, and non-binary people, not just cisgender people, in the studies investigating robot's genderedness.}
\end{quote}
\vspace{0.1cm}

This guideline also urges to drop the biologized and essentialist way of asking about sex on a female/male categorical binary. The distinction of sex/gender and the deterministic understanding of sex as a binary biology is highly criticized within the neuro- and biofeminist field \cite{bluhm2012neurofeminism}. Instead, understanding the terminology of the variety of gender identities that are actually relevant for social interaction as well as actively employing diverse recruiting efforts are needed. Scheuerman et al. drafted a living document \href{https://www.morgan-klaus.com/gender-guidelines.html}{``HCI Guidelines for Gender Equity and Inclusivity"} containing a section on gender inclusive research methods which gives valuable insights into how to perform inclusive research. For instance, they suggest using the following options to ask about participants' gender: woman, man, non-binary, prefer not to disclose, prefer to self-describe and explain how to carry out in-person studies in a way that is respectful of all gender identities (see also \cite{spiel2019better}).

The studies we reviewed often lacked a clarifying definition of ``gender''. Only Bryant et al. \cite{bryant2020should} attempted a description of the gendering process as related to robots. We do not advocate for a universal fixed definition of gender that could fit all research and researchers. However, we think it is important for researchers working on this topic to:

\begin{quote}
    \textbf{Guideline 2}: \textit{Reflect on their understanding and experience of ``gender," clarify this understanding in their paper, and explain the reason why they are gendering the robot in their study.}
\end{quote}
\vspace{0.1cm}

A practical way forward to fulfill this objective is to go through a self-assessment process where the researcher(s) ask(s) themselves: (i) What does gender mean to me? (ii) Is gendering really needed to answer my research question? (iii) Am I embedding gender stereotypes in the study? We argue that by making the gendering process a reflective part of designing a study (as suggested by Weiss and Spiel \cite{weiss2021should}) and especially visible in writing about a study, most of the stereotypes imbued into gendered robots might be avoided. 

Another methodological pitfall we observed in some of the studies, which is unfortunately endemic to HRI research, is the uniformity of participants' characteristics. Most of the reviewed studies resorted to a sample of young participants (possibly university students). The main drawback of the homogeneity in participants' characteristics is that it makes it difficult to address context- and user-specific differences. We acknowledge that resorting to students as participants is oftentimes dictated by the research complexity level or by the lack of funding to recruit a more diverse set of participants. However, in the specific context of gendering robots, this might give one-sided results, as individual participants' characteristics might disclose relevant insights into how gendered robots are perceived. While we put forward a caveat in this sense, we do not feel like enforcing a guideline, as the use of university students as participants might depend on the economic availability of each research group. 

From a methodological perspective, we need to mention another aspect we observed in the reviewed studies, which might constitute a limitation of this review, namely the richness of robots, tasks, and activities. The studies we reviewed used many different robotic platforms and envisioned many different tasks (e.g., observing pictures, watching videos, interacting with the robot), activities (e.g., educational activities, casual conversations) and participants' roles (e.g., remote observer, co-present observer, interactant). This complexity is not bad in principle, but is risky when building a research field from scratch as it makes comparability between studies difficult, thus hindering the possibility of drawing conclusions on the role of robot's genderedness as a whole. To circumvent this, we suggest to:
\vspace{0.1cm}

\begin{quote}
\textbf{Guideline 3}: \textit{Focus on few application scenarios (e.g., healthcare, education, hospitality) and perform studies under comparable conditions.}
\end{quote}
\vspace{0.1cm}

This way the HRI scholarship could adopt an incremental approach to the study of robot's genderedness, where scientific clarity is prioritized over novelty, and in turn encourage replication studies where existing experimental designs are reused with slightly different variables to check if results still hold.

\subsection{Discussion on Manipulation of Robot's Genderedness (RQ1)}

Through this scoping review, we discovered that the robot’s genderedness has been manipulated by the HRI scholarship using cues such as the robot's voice, name, facial features, apparel, colors, body proportions, and hairstyle. Some of these cues are fruit of social conventions and socio-cultural schemata (e.g., names, hairstyle, apparel), others refer to the physical and physiological characteristics of gendered bodies (e.g., the waist-to-hips ratio and the voice frequency). Nevertheless, most of them tap into a binary understanding of gender. Indeed, in 89\% of the reviewed studies, the gendering of the robot has been manipulated within the female/male binary. 
As a result, we draw the following guideline:
\vspace{0.1cm}

\begin{quote}
\textbf{Guideline 4}: \textit{Avoid imbuing robots with oversimplified and normative visions of gender as binary.}
\end{quote}
\vspace{0.1cm}

One way to go about this objective is for researchers to engage in a critical reflection of their own gendering process by asking themselves: (i) Are the gender cues I have chosen really needed? (ii) Can I achieve the manipulation of genderdness with less and more subtle cues? (iii) Why am I manipulating the robot's genderedness with these cues? (iv) Am I embedding gender stereotypes in the robot by using these cues?  
Since gender cues might layer and affect each other in unexpected ways, it might also be a good strategy to either choose a robot with quite an undefined gender attribution at baseline and add additional gender cues to it or investigate the robotic embodiment for its existent genderedness without manipulating its design. 
Tools like the humanoid ROBOts - Gender and Age Perception (ROBO-GAP) dataset \url{https://robo-gap.unisi.it/} could help researchers choosing the right robot and checking its perceived gender already at baseline.
This brings us to the fifth guideline:
\vspace{0.1cm}

\begin{quote}
\textbf{Guideline 5}: \textit{Perform a pre-test of the genderedness of the robotic platform you plan to use to avoid further gendering when it is not needed.} 
\end{quote}
\vspace{0.1cm}

As a non-negligible aspect of the gendering process observed in the reviewed studies, most of the gender cues were used in combination with others and only rarely in isolation, as if the layering of these cues could strengthen the gender attribution. However, from the results of the manipulation check, it becomes apparent that gender is attributed to robots on the basis of the tiniest gender cues (see Rea et al. \cite{rea2015check}). Besides, overdoing gender cues and/or using extremely stereotypical cues (e.g., pink ribbon/ blue bow-tie) might lead to stronger stereotyping \cite{perugia2023models} and end up revealing the purpose of the study. Since the layering of gender cues does not yield any additional effect on the manipulation of gender and also puts researchers at risk of stereotyping, we strongly recommend to:
\vspace{0.1cm}

\begin{quote}
    \textbf{Guideline 6}: \textit{Avoid stereotypical gender cues and use as little gender cues as possible, and as subtle gender cues as feasible.}
\end{quote}
\vspace{0.1cm}
 
Even though most reviewed studies presented the robots through a physical embodiment, the context(s) in which the robots were shown varied widely. The process of gendering is not just initiated by the presence of certain appearance cues, but is deeply influenced by the context where the interaction takes place. Interacting with robots that have a certain role is different than attributing gender to a robot in a contextless task \cite{perugia2022shape}. It is \textit{during} the interaction that the most performative aspects of the robot's genderedness unfold and become apparent and it is \textit{through} the interaction that the robot's genderedness acquires a symbolic meaning 
\cite{bisconti2021we}. Hence, we recommend researchers to:
\vspace{0.1cm}
\begin{quote}
\textbf{Guideline 7}: \textit{Consider the interaction context as part of the manipulation of the robot's genderedness and study how the gendering of robot's roles, behaviors, and activities influences the gender attribution to the robot or even flips it.}
\end{quote}
\vspace{0.1cm}

Another striking result of this scoping review was that almost half of the studies did not perform any statistical analysis to assess whether the manipulation of the robot's genderedness actually succeeded. This is particularly problematic as it makes it difficult to establish whether the lack of significant main effects of the robot's genderedness on the dependent variables is actually due to robot's genderedness or to its unsuccessful manipulation. Future studies should: 
\vspace{0.1cm}

\begin{quote}\textbf{Guideline 8}: \textit{Always perform a manipulation check to test whether the robot's genderedness is perceived by participants in the expected way.}
\end{quote}
\vspace{0.1cm}

In Ghazali et al. \cite{ghazali2018effects}, the manipulation check was deemed successful since the female and male robot conditions were perceived as significantly different in terms of gender. When taking a look at the descriptive statistics reported by the authors, however, one can notice that the robot's perceived genderedness did not differ in terms of categorization. 
Based on this, we recommend researchers to perform a manipulation check, but also:
\vspace{0.1cm}

\begin{quote}
    \textbf{Guideline 9}: \textit{Check the descriptive statistics of each gender condition as part of the manipulation check, as a significant difference between conditions does not necessarily grant a different categorization of the robot's genderedness.}
\end{quote}
\vspace{0.1cm}

Measuring the robots' genderedness is not exempt from shortcomings. A research concept is necessarily entangled with the questionnaire that asks the participant about it \cite{lisy2019sexual}. Meaning, if the concept is a binary understanding of gender, then a question about feminine or masculine aspects whether in one or different items, will ontologically reproduce a binary idea of gender. Besides, asking people to attribute gender to a robot might result in a gender attribution even when the robot is not perceived as gendered in the first place. In this scoping review, we identified several quantitative ways to measure the robot's genderedness. However, it might be interesting to:

\vspace{0.1cm}

\begin{quote}\textbf{Guideline 10}: \textit{Explore more subtle ways of checking whether gender is attributed to the robot, for instance, through qualitative or indirect measures.}
\end{quote}
\vspace{0.1cm}

For instance, Roesler et al. \cite{roesler2022context} used naming frequency to understand how the robots in their study were attributed a gender, which gave participants the possibility not just to give robots traditional names, but also technical and more object-oriented ones.

\subsection{Discussion on Effects of Robot's Genderedness on Perceptions of and Interactions with Robots (RQ2)}

When taking the results as whole, it becomes quite clear that gendering robots has a strong effect on stereotyping. 
We cannot help but wonder whether the effects that robot's genderedness has on stereotyping might have been due to the way the robot was gendered in the first place. As to say, if we imbue robots with stereotypical gender cues, it might become difficult for participants to not stereotype them as a result.

In general, one of the clear-cut outcomes of this scoping review is that genderedness does not have an effect on crucial constructs for the HRI, such as acceptance and likability, as it perhaps does for voice assistants. In this regard, however, the studies published in the last year paint a different picture. They disclose that in service contexts, female robots and gender ``congruity" (i.e., the match between participant's gender and robot's genderedness) are almost always preferred. Comparing these results with the research on voice assistants, it seems that there is something in the service context that makes the female genderedness of artificial agents immediately relevant. As if the fact that we as humans are used to see women in service roles makes the suitability of female robots in the same role immediately glaring. From a feminist standpoint, a question arises: do we have to second the preference of the user for female service robots even if we know it stems from a discriminatory understanding of a gendered society? We as authors argue that we do not have to, and present the HRI community with a guideline that could serve as a design opportunity:
\vspace{0.1cm}

\begin{quote}\textbf{Guideline 11}: \textit{Use gendered robots to offer occasions of defamiliarization with normative gender roles and disrupt binary conceptualizations of human gender and tasks.}
\end{quote}
\vspace{0.1cm}

In the context of interaction effects, two results caught our attention in the papers we reviewed. Calvo-Barajas et al. \cite{calvo2020effects} discovered that children perceived a female robot as less likable when it expresses high anger instead of more positive or less intense emotions, while Jackson et al. \cite{jackson2020exploring} disclosed that male participants like male robots but not female robots when they issue strong rejections. These results seem to suggest that female robots, like women, are liked less when they are not compliant or not consensual. This follows the problematic narrative that wants women submissive and aware of ``their place" in the world. 
In a real-life environment, how should a female robot react to people issuing annoyance for their lack of compliance or consent? Should they maintain a jokey vibe of servitude as voice assistants originally did \cite{west2019blush} or react resolutely as in Winkle et al. \cite{winkle2021boosting}? We consider Winkle et al.'s work \cite{winkle2021boosting} a valid and viable option. Aside from this, however, the HRI scholarship should start reflecting on the ethical implications of gendered robots and their (mis)treatment, especially given the highly symbolic meaning human-human\textit{oid} interactions entertain with human-human interactions \cite{sparrow2020robotics,sparrow2017robots,perugia2022shape,zhao2006humanoid}. As such we suggest a last guideline:
\vspace{0.1cm}

\begin{quote}\textbf{Guideline 12}: \textit{Critically reflect on the results of your research on gendered robots and engage with a discussion of the ethical implications of your findings, especially considering the highly symbolic value of human-humanoid interactions for human-human relations.}
\end{quote}
\vspace{0.1cm}

For future robot designs, the challenge remains whether we could come close to a gender neutral or even genderless humanoid robot. Since the human form is so strongly interconnected with the gendering process \cite{perugia2022shape}, the predominant use of a humanoid design form could be put into question. 
The HRI scholarship might want to identify alternatives to humanoid designs as well as imagine interactions with robots that do not just mimic human-human interactions.

\vspace{0.3cm}

\textbf{Authors' Contributions.}
GP formulated the research questions, devised the inclusion and exclusion criteria, performed the string search, went through the selection pipeline, read all the papers, extracted the information from the papers, summarized all the results, wrote sections 2, 3, 4, 5, and 6 of the paper, contributed to the writing of section 1, and prepared all the tables. DL formulated the research questions, devised the inclusion and exclusion criteria, performed the string search, read part of the papers, extracted information from part of the papers, wrote section 1 of the paper and contributed to the writing of section 6.\vspace{0.3cm}

\textbf{Acknowledgements.}
The authors would like to thank Laura van der Bij for performing the strings search for the papers in the addendum, and Latisha Boor for checking the correctness of Tables 1 and 2, contributing to the creation of Table 4, and designing the visualizations in Figures 2, 3, 4, and 5. 
\vspace{0.3cm}

\textbf{Funding}
GP's work is partly funded by the research programme Ethics of Socially Disruptive Technologies (ESDiT), which is funded through the Gravitation programme of the Dutch Ministry of Education, Culture, and Science and the Netherlands Organization for Scientific Research (NWO grant number 024.004.031). DL's work is funded by the Wallenberg AI, Autonomous
Systems and Software Program – Humanities and Society (WASP-HS) project ``The Ethics and Social consequences of AI and caring Robots.'' 
\vspace{0.3cm}

\textbf{Conflict of Interest}
The authors declare that they have no conflict of interest or competing financial interests or personal relationships that could have appeared to
influence the work reported in this paper.

\setlength{\LTcapwidth}{21.2cm}
\clearpage
\onecolumn
\begin{landscape}
\footnotesize
\arraystretch{2}
\footnotesize\begin{longtable}{ >{\raggedright\arraybackslash}p{2.5cm} | p{0.5cm}  >{\raggedright\arraybackslash}p{5.2cm} | >{\raggedright\arraybackslash}p{4.5cm} | >{\raggedright\arraybackslash}p{8.5cm}}
\caption{Experimental information about the studies included in the scoping review: Authors (Date), independent variables (\textit{bs}= between subjects; \textit{ws}= within subjects), dependent variables (in \textbf{bold}, the significant main effects of robot's genderedness on the dependent variables, i.e., $p<.05$), and summary of findings.} \label{table:Experimental_Information} \\
     & \multicolumn{2}{c}{\textbf{Variables}} & \\
\textbf{Author (date)} & \multicolumn{2}{l|}{\textbf{Independent}} & \textbf{Dependent*} & \textbf{Summary}\\
\hline\noalign{\smallskip}
Bernotat et al. & bs & \textit{Robot Genderedness} (female, male) & communion, agency, & The female robot was perceived as more suitable \\
(2017) \cite{bernotat2017shape} & & & \textbf{affective trust}, \textbf{cognitive} & for stereotypically female tasks and induced more\\
& & & \textbf{trust, task preference} & affective and cognitive trust than the male robot.\\[2ex]
Bernotat et al. & bs & \textit{Robot Genderedness} (female, male) & \textbf{communion}, agency, & The female robot was perceived as more communal\\
(2021) \cite{bernotat2021fe} & & & \textbf{affective trust}, \textbf{cognitive}, & and suitable  for stereotypically female tasks and induced\\
& & & \textbf{trust, task preference} &  more affective and cognitive trust than the male robot.\\[2ex]
Bryant et al. & bs & \textit{Robot Genderedness} & occupational competency, & The perceived gender of the robot did not have an \\
(2020) \cite{bryant2020should} & & (female, male, neutral) & trust in occupational & impact on perception of occupational competency \\
& & &  competency & nor on trust in the robot's occupational competency.\\[2ex]
Calvo-Barajas & ws & \textit{Robot Genderedness} (female, male) & competence & The robot's genderedness did not influence the  \\
et al. (2020) \cite{calvo2020effects}& bs & Emotion Type (happiness, anger) & (smartness, helpfulness), & perception of competence, trust, and likability \\
& bs & Emotion Intensity & likability (appearance, & of the robot. However, it interacted with emotion \\
& & (low, medium, high) & & intensity to affect the
robot’s perceived likability \\
&&&& (apperance).\\[2ex]
Chita-Tegmark & bs &  \textit{Agent Genderedness} (female, male) & \textbf{emotional intelligence} & Male agents (human/robot) were perceived to have \\
et al. (2019) \cite{chita2019gender}& bs &  Agent Type (human, robot) & & higher levels of emotional intelligence (EI) than female \\ 
& bs & Participant Gender (female, male)  & & agents. Female participants perceived female agents as \\
& bs & Emotional Intelligence (low, high) & & having lower EI than male agents in the low EI condition.\\[2ex]
Eyssel \& Hegel & ws & \textit{Robot Genderedness} (female, male) & \textbf{communion}, \textbf{agency}, & Participants ascribed more agentic traits to the male \\
(2012) \cite{eyssel2012s} &  &  & \textbf{preference as task} & robot and more communal traits to the female robot. \\
& & & \textbf{interaction partner} &  The male robot was rated as more suitable for stereo-\\
& & & & typical male tasks and the female robot was evaluated \\
& & & & as more suitable for stereotypically female tasks. \\[2ex]
Eyssel et al. & bs & \textit{Robot Genderedness} (female, male) & anthropomorphic inferences & Participants revealed greater HRI acceptance and \\
(2012a) \cite{eyssel2012if} & bs &  Participant Gender (female, male) & (attribution of mind), & felt more psychological closeness for the same-gender \\
& bs & Voice Type (humanlike, robotlike) & psychological closeness, & robot compared to the opposite-gender robot. \\
& & & acceptance & Participants anthropomorphized the same-gender  \\
& & & & robot more strongly than the opposite-gender robot.\\[2ex]
Eyssel et al. & bs & \textit{Robot Genderedness} (female, male) & likability (warmheartedness) & Participants perceived the same-gender robot as more\\
(2012b) \cite{eyssel2012activating}& bs & Participant Gender (female, male) & anthropomorphism (human & likable and psychologically close than the opposite- \\
& bs & Voice Type (humanlike, robotlike) & essence), contact intentions & gender robot.\\[2ex]
Ghazali et al.  & bs & Human-\textit{Robot Gender} Match  & perceived intelligence, & Psychological reactance was lower towards the same- \\ 
(2018) \cite{ghazali2018effects} & & (mismatch, match) & anthropomorphism, trust &gender robots compared to the opposite-gender \\
& bs & Trust Face & beliefs (trust, perceived trust, & robots.\\
& & (least, most trustworthy) & individualized trust), trusting & \\
& & & behaviors, feeling of anger, & \\
& & & negative cognitions & \\[2ex]
Jackson et al. & bs & \textit{Robot Gender} (female, male) & politeness, harshness, & There was no significant main effect of the robot's \\ 
(2020) \cite{jackson2020exploring} & bs & Participant Genderedness (female, male) & directness, likability & genderedness on the dependent variables, but many   \\
& bs &  Human Requester Gender & & significant interaction effects between robot's   \\
& & (female, male) & & genderedness and the other independent variables. \\ 
& ws & Severity Moral Infraction  & &  These interaction effects seemed to point to the  \\
& &(low, high) & & fact that male robots are perceived more favorably \\ 
& ws & Face Threat Robot Response  & & than female robots when rejecting commands.\\
& & (low,high) & & \\[2ex]
Jung et al. & bs & \textit{Robot Genderedness} (female, male) & \textbf{anthropomorphism}, & The male robot was perceived as more anthropomorphic,\\
(2016) \cite{jung2016feminizing} & bs & Cue Location (body, screen) & \textbf{anxiety}, \textbf{animacy} &more animate and less anxious than the female robot\\
& & & &regardless of the gender cue location. \\[2ex]
Kraus et al. & bs & \textit{Robot Explicit Genderedness} & competence, likability, trust, & The male personality was perceived as significantly \\
(2019) \cite{kraus2018effects} & & (female, male) &  reliability, acceptance, &  more trustworthy and predictable in the taxi ordering\\ 
& bs &  Robot Personality (female, male) & predictability & scenario. The female personality was perceived as more\\
& ws & Task Scenario (female, male) & &  likable in both scenarios.\\[2ex]
Kuchenbrandt  & bs & \textit{Robot Genderedness} (female, male) & task-related competence, & In the female task context, participants made more  \\ 
et al. (2014) \cite{kuchenbrandt2014keep}& bs & Gender Typicality Task  & follow directions, acceptance, & errors, were less willing to accept help from the robot\\
& &(female, male) & speed of completion, errors &   on a future task, and anthropomorphized the robot to \\
& bs & Participant Gender (female, male) & & a smaller degree than in the male task.\\[2ex]
Law et al. & bs & \textit{Robot Genderedness} (female, male) & \textbf{trust} & Robots in the high emotional intelligence (EI) condition  \\ 
(2020) \cite{law2021interplay} & bs & Emotional Intelligence (low, high) & & were trusted more than the robots in the low EI condition. \\ 
& bs & Presentation Style (text, voice) & &  Male robots were trusted more than female robots.\\[2ex]
Law et al. & bs & \textit{Robot Genderedness} (female, male) & trust & The robot's genderedness did not affect participants'\\
(2020) \cite{law2021interplay} & bs & Performance-based Trust & & trust. \\
& & (low, high) & &  \\
& bs & Presentation Style (text, voice) & \\[2ex]
Lugrin et al. & bs & \textit{Robot Genderedness} (female, male) & perceived competence & The robot's genderedeness did not affect people's  \\ 
(2020) \cite{lugrin2020if} & bs & Language Type  & likability, social skills & perception of the robot's competence, likability and \\
& & (standard, accent, dialect) & &social skills.\\[2ex]
Makenova et al. & bs & \textit{Robot Genderedness} (female, male) & trust, \textbf{donations}, & The female robot received more donations from \\ 
(2018) \cite{makenova2018exploring} & bs & Participant Gender (female, male) & credibility, engagement & participants compared to the male robot.\\[2ex]
Nomura \&  & bs & \textit{Robot Genderedness} (female, male) & 20 pairs of adjectives, only & The female robot was perceived as prettier and more  \\
Kinoshita &  & & 4 specified: \textbf{pretty-hateful}, & familiar than the male robot. \\
(2015) \cite{nomura2015gender}* &  & & \textbf{familiar-unfamiliar}, mild & \\
&&& -terrible, pleasant–unpleasant &\\[2ex]
Nomura \& & bs & \textit{Robot Genderedness} (female, male) & politeness, mildness, & The robot's genderedness did not have a significant \\ 
Takagi & bs & Participant Gender (female, male) & ambitiousness, assertiveness & effect on politeness, mildness, ambitiousness, assertiveness. \\ 
(2011) \cite{nomura2011exploring} & bs & Participant Background & & \\
& & (tech., social sciences) & & \\[2ex] 
Paetzel et al.  & bs & \textit{Robot Genderedness} (female, male) & intelligence, likability, & The male robot was perceived as less anthropomorphic \\ 
(2016a) \cite{paetzel2016congruency} & bs & Modality (unimodal, multimodal) & uncanniness, &  than the female robot, and especially so in the multimodal \\
& & & \textbf{anthropomorphism}, trust, & condition. The female robot was perceived as more intelligent,  \\
& & & relaxation, contentment & likable for entertaining a conversation, pleasant, relaxed and  \\
&&&& content than the male robot in the multimodal condition.\\ 
&&&&On the opposite, the male robot was perceived as significantly\\ 
&&&&  more familiar and trustworhty than the female robot in the\\
&&&&  unimodal condition. \\ [2ex]
Paetzel et al. & bs & \textit{Genderedness Robot} Face & likability, uncanniness, & The robot's genderedness did not affect the perception  \\ 
(2016b) \cite{paetzel2016effects} & bs &  (female, male)  & anthropomorphism, \textbf{trust} & of pleasantness (i.e., likability), strangeness (i.e., uncan-\\
&  & \textit{Genderedness Robot} Voice   &  &  niness), anthropomorphism. The congruent female robot \\
& & (female,male)& &   was perceived as more trustworthy than the congruent\\
&&&&  and incongruent male robots.\\[2ex]
Pfeifer \& Lugrin & bs & \textit{Robot Genderedness} (female, male) & objective learning success & Robot's genderedness did not have any significant \\ 
(2018) \cite{pfeifer2018female}& bs & Learning Material (female, male) & &  main effect on learning success. Participants who took  \\
& & & &part in the learning activity with a female robot \\
&&&& explaining male learning materials obtained the\\
&&&&highest learning success.\\[2ex] 
Powers \& & bs & \textit{Robot} Voice \textit{Genderedness}  & following robot's advice \textit{(not} & 100\% of the participants who saw the robot with a short 
 \\
Kiesler & &(female, male, female & \textit{the only dependent variable } & chin (childlike) and heard the robot speak in the undam-\\ 
(2006) \cite{powers2006advisor} & & dampened, male dampened) & \textit{but the only one used to test } & pened male voice said they would take the robot’s advice. \\ 
& bs & chin height (short, long) & \textit{differences across conditions)} &When the chin was long (adult) and had a male voice, or the \\
& bs & forehead height (short, long) & & voice was female and the chin was short, the percent went down \\
&&&& to 91\%. only 50\% of the participants who saw the robot with \\
&&&& a long chin and heard it speak with a female voice said they \\
&&&&would take its advice. \\[2ex] 
Powers et al. & bs & \textit{Robot Genderedness} (female, male) & compassion, dominance, & Participants said more words to the male robot than to  \\ 
(2005) \cite{powers2005eliciting} & bs & Participant Gender (female, male) & knowledge, likability, & the female robot. Men uttered more words than women in  \\ 
& ws & Target of Conversation  & personality, \textbf{spoken words} &  the female robot condition, whereas women said more words  \\
& & (female, male) & & than men in the male robot condition.\\ [2ex]
Rea et al. & bs & \textit{Robot Genderedness} (female, male) & communion, politeness, & Female participants showed more politeness towards the \\ 
(2015) \cite{rea2015check} & bs & Participant Gender (female, male) & agency, engagement, task & robot than male participants. Participants did not apply \\
&  & & preference, relaxation & gender stereotypes to the robots based on their gendered-\\
&&&&ness. The robot's genderedness did not affect engagement.\\[2ex]
Reich-Stiebert  & bs & \textit{Robot Genderedness} (female, male) & warmth, agency, competence, & Participants showed not to apply gender stereotypes to \\ 
\& Eyssel & bs & Participant Gender (female, male) & willingness to engage, obj. & robots when interacting in gender stereotypical tasks. A \\ 
(2017) \cite{reich2017ir} & bs & Task Gender Typicality  & learning outcome, subj. & mismatch between robot's genderedness and gender typicality  \\
& & (female, male) & learning outcome, learning & of the task led to increased willingness for future learning with \\
& & & mood, intrinsic motivation, & the robot.\\
& & & perception as learning & \\
& & & companion, satisfaction & \\[2ex]
Sandygulova  & ws & \textit{Robot} Gendered \textit{Voice}   & likability, robot type, & Young children preferred to interact with a same-gender \\ 
\& O'Hare & & (female, male)  & robot as social actor, mood, & robot compared to a opposite-gender robot, whereas \\ 
(2018) \cite{sandygulova2018age} & bs & Children Gender (female, male) & valenced affective state, & older children reported no difference in their preference.  \\
& bs & Children Age (5-8 yo, 9-12 yo) & happiness, \textbf{smiling} & Children smiled more during the interaction with the \\
&&&&female robot than during the interaction with the male\\
&&&&robot.\\[2ex]
Sandygulova  & bs & Human-\textit{Robot Gender Match} & duration of interaction, & Children spent more time with the same-gender \\
\& O'Hare  &&(mismatch, match) & proximity & robot and their proximity with the robot decreased or \\
(2016) \cite{sandygulova2016investigating}& & & &  remained stable when interacting with a male robot and\\
&&&&increased when interacting with a female robot.\\[2ex]
Sandygulova & ws & \textit{Robot} Voice \textit{Genderedness} (female UK, & voice preference & Children preferred the robot with the English UK accent \\
\& O'Hare & & male UK, female US, male US) & & over the robot with the English US accent. No effect of \\
(2015) \cite{sandygulova2015children} & & & & gender on voice preference is reported.\\[2ex]
Sandygulova & bs & \textit{Robot} Voice \textit{Genderedness} and Age & happiness & Female children expressed more happiness towards the \\
et al. (2014) \cite{sandygulova2014investigating} &  & (female child, male child,   & & female robot. Male children expressed more happiness \\
& & female adult) & & towards the male robot\\
& bs & Children Age (8, 9, 10, 11, 12 yo) & & \\[2ex]
Siegel et al.  & bs & \textit{Robot Genderedness} (female, male) & trust, \textbf{donations}, credibility, & Men trusted the female robot more, donated more money 
 \\
(2009) \cite{siegel2009persuasive} & bs & Participant Gender (female, male) & engagement &to it, considered it more credible, and felt more engagement\\
& bs & Participant alone (not alone, alone)  & &  with it. Women showed no preference for any of the robots in terms of trust and donations, but rated the male robot as more \\
& & & & credible and more engaging than the female robot.\\[2ex]
Tay et al. & bs & \textit{Robot genderedness} (female, male) & perceived trust, acceptance, & The perceived gender of the robot did not affect perceived \\ 
(2014) \cite{tay2014stereotypes} & bs & Robot Personality  & attitude towards robots, & trust, acceptance, attitude towards the robot, subjective\\ 
& & (introverted, extroverted) & subjective norms, affective &  norms, cognitive evaluations. Participants showed higher \\
& bs & Occupational role (female, male) & evaluation, cognitive evaluation, &affective evaluations and perceived behavioral control in the \\
&  & & perceived behavioral control& female healthcare robot condition, and higher affective  \\
& & &  & evaluations in the male security robot condition. Some \\
&&&&of the non-significant results were marginally significant.\\[2ex]
Thellman & bs & \textit{Robot Genderedness} (female, male) & persuasiveness & The robot's genderedness did not have a significant effect\\ 
et al. (2018) \cite{thellman2018he}& bs & Participant Gender (female, male) & & on persuasiveness, nor did participants' gender.\\ [2ex]
You \& Lin  & bs & \textit{Robot} Voice \textit{Genderedness}  & trust, donations, credibility & The genderedness of the robot did not affect trust,  \\ 
(2019) \cite{you2019gendered} & &(female, male) & engagement & the amount of donations, credibility, and engagement.  \\
& bs &\textit{Robot} Appearance \textit{Genderedness}& & While men rated the female robot as more credible (safety) \\
& &  (female, male, neutral) & &  than the male robot, women did not show any difference  \\
& & & & in their ratings. \\[2ex]
Zhumabekova et & ws &\textit{Robot Genderedness} (female, male) & liked interacting with robot & Children liked the interaction with a same-gender robot  \\ 
 al. (2018) \cite{zhumabekova2018exploring} & bs & Participant Gender (female, male) & & more than the interaction with a robot with a different\\
 &&&&gender.\\[2ex] 
Steinhaeusser et  & bs & \textit{Robot} Voice \textit{Genderedness}  & anthropomorphism, & There was no significant effect of robot's genderedness  \\ 
al. (2021) \cite{steinhaeusser2021anthropomorphize} &  & (female, male, neutral)&  transportation, attitudes & on anthropomorphism, the attitude towards robots, and \\
& bs & Participant Gender (female, male) & towards robots & the transportation participants felt when the robot told  \\
& & & &the story, nor were there any interaction effects between\\
&&&&participant's gender and robot's genderedness \\[2ex]
\hline
\end{longtable}

\end{landscape}
\clearpage
\twocolumn

\newpage

%



%
%

\bibliographystyle{spmpsci}      
\bibliography{references.bib}  

%
%

\end{document}